\documentclass{article}

    \PassOptionsToPackage{numbers, compress}{natbib}
 \usepackage[final,main]{neurips_2026}

\usepackage[utf8]{inputenc} 
\usepackage[T1]{fontenc}    
\usepackage{hyperref}       
\usepackage{url}            
\usepackage{booktabs}       
\usepackage{amsfonts}       
\usepackage{nicefrac}       
\usepackage{microtype}      
\usepackage{xcolor}         
\usepackage{wrapfig}
\usepackage{amsmath} 
\usepackage{multirow}
\usepackage{microtype}
\usepackage{graphicx}
\usepackage{subfigure}
\usepackage{booktabs} 
\usepackage[ruled,vlined,linesnumbered]{algorithm2e}
\usepackage{colortbl}

\usepackage[utf8]{inputenc} 
\usepackage[T1]{fontenc}    
\usepackage{hyperref}       
\usepackage{url}            
\usepackage{booktabs}       
\usepackage{amsfonts}       
\usepackage{nicefrac}       
\usepackage{microtype}      
\usepackage{xcolor}         
\usepackage{colortbl}
\definecolor{customcolor}{HTML}{D3D3D3}
\usepackage{algorithmic}
\title{Detect What You Need: Chain-of-Causal Reasoning for 3D Intent Grounding}

\author{
Zihao Zhang\textsuperscript{1} \quad
Aming Wu\textsuperscript{2} \quad
Yang Li\textsuperscript{1} \quad
Yahong Han\textsuperscript{1}\thanks{Corresponding author.}
\\[0.6em]
\textsuperscript{1} School of Artificial Intelligence, College of Intelligence and Computing, Tianjin University, China
\\
\textsuperscript{2}School of Computer Science and Information Engineering, Hefei University of Technology, China
\\[0.4em]
{\ttfamily\small
\{zhangzihao2490, liyang1389, yahong\}@tju.edu.cn
}
\\
{\ttfamily\small amwu@hfut.edu.cn}
}

\begin{document}

\maketitle

\vspace{-15pt}
\begin{abstract}
Accurately matching human intentions in 3D space is an important goal of artificial intelligence. Recently, 3D Intention Grounding (3D-IG) has emerged, aiming to localize target 3D objects given a natural-language intent. Unlike conventional visual grounding with descriptive referring expressions, 3D-IG intents are abstract and non-descriptive, making object localization substantially more challenging. This requires inferring latent functional requirements from non-descriptive intents and aligning them with object-level 3D representations. However, existing methods largely rely on implicit intent–object matching, leading to logical gaps and limited interpretability, robustness, and generalization. To address these challenges, we propose Chain-of-Causal Reasoning (CoCR), a causality-inspired functional dependency reasoning framework that explicitly bridges abstract intentions and candidate objects through intermediate functional requirements. Specifically, CoCR progressively decomposes complex intents into ordered functional requirements, thereby forming an explicit intent–function–object reasoning chain that links abstract intentions to object suitability. 
Building on this chain, we construct a causality-inspired functional dependency graph to model requirement--attribute relationships and introduce a causal-visual alignment module that aligns function-aware representations with the geometric-semantic features of 3D point clouds, enabling bidirectional verification between structured functional reasoning and visual evidence.
Extensive experiments on 3D Intention Grounding and 3D Visual Grounding demonstrate that our method enhances intent-aware object localization. Our code is available at \url{https://2490o.github.io/CoCR}.
\end{abstract}    
\vspace{-5pt}
\section{Introduction}
\vspace{-5pt}
\label{sec:intro}

In real-world human-computer interaction scenarios, particularly in applications involving embodied intelligence \cite{bip3D, embodiedscan}, humans typically express their needs through natural language (e.g., ‘I need back support’), rather than providing explicit object references \cite{VPPNet,vg2} (e.g., ‘the pillow on the sofa’). Therefore, compared to the traditional 3D Visual Grounding task \cite{vg2}, the recently proposed 3D Intent Grounding (3D-IG) task \cite{intent3d} is more suitable for real-world applications. The core objective of this task is to enable models to interpret a natural-language intent, infer the functional requirements, and then accurately match and localize the corresponding target objects in 3D scenes.

\begin{figure}[!t]
	\centering
        
	\includegraphics[width=1.0\columnwidth]{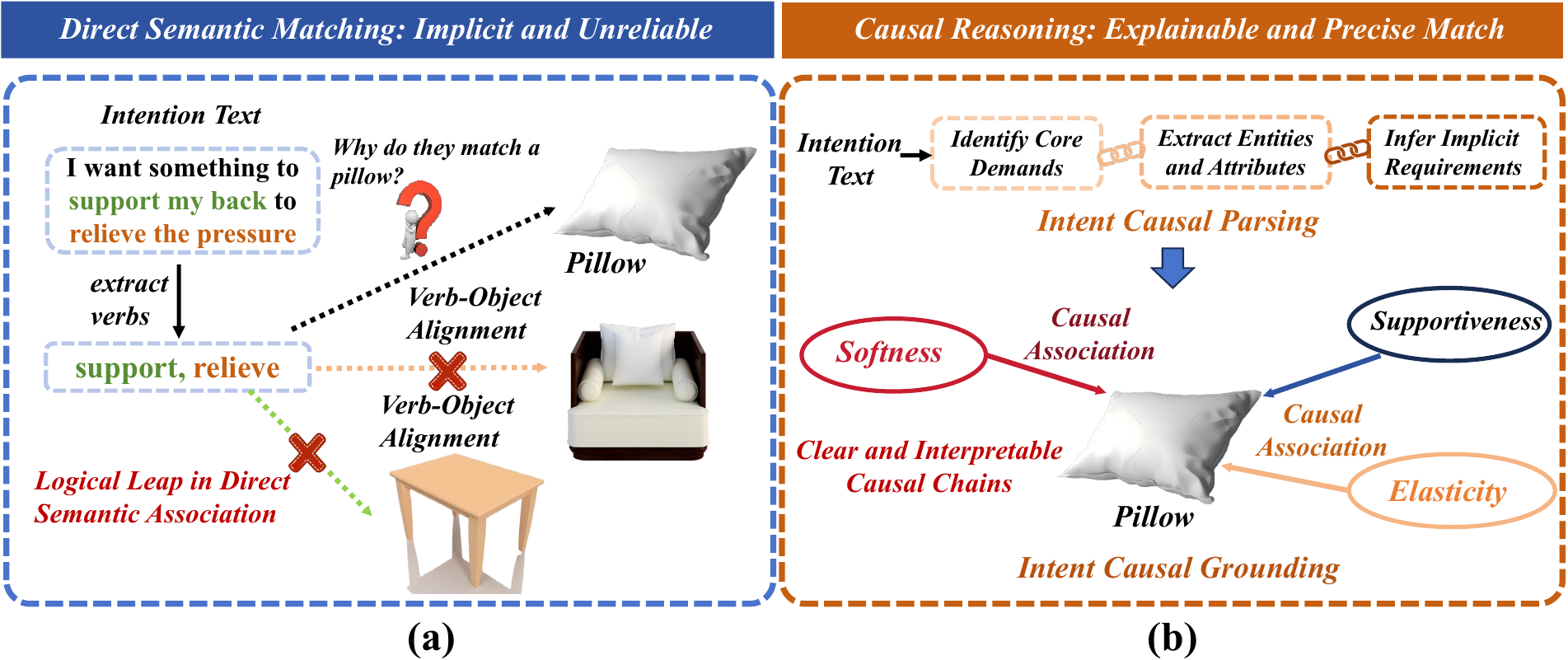}
	\DeclareGraphicsExtensions.
	\begin{center}
		\vspace{-10pt}
		\caption{(a) Existing methods rely on direct semantic matching, leading to logical leaps. Our method performs step-by-step functional causal reasoning by explicitly parsing intents into functional requirements. (b) The workflow of Chain-of-Causal Reasoning builds explicit intent–function–object dependency paths for object reasoning and localization.
        }\label{f1}
	\end{center}
	\vspace{-20pt}
\end{figure}

The core challenge of the 3D-IG task lies in the abstract nature of human intent. Unlike the traditional Visual Grounding task, which relies on explicit object categories or visual descriptions, the intent texts processed by 3D-IG do not point to specific object characteristics. Instead, they focus on functional attributes and usage contexts. This requires the model first to infer implicit functional requirements from vague expressions and then match them with 3D object features, significantly increasing the complexity of the task. Existing methods \cite{intent3d} match target objects by establishing semantic correspondences between ‘verbs and objects,’ relying on direct alignment between verbs and objects, as shown on the left side of Figure \ref{f1}. However, this direct semantic association introduces a clear logical gap, as the model does not understand why a particular object should match a given intent. Specifically, the model can only match at the semantic level but fails to infer whether an object truly satisfies the implicit functional requirements of the intent. For example, when faced with the intent ‘I need something to support my back,’ the model may match ‘support’ and ‘back’ with multiple objects, such as a sofa, pillow, or chair, but it cannot determine which object actually provides ‘back support,’ leading to potential mismatches and incorrect localization in open-world scenarios. Therefore, existing methods often lack an explicit modeling of the functional dependencies between intents and object suitability, limiting their robustness and generalization.

To address these issues, we propose a new method, i.e., Chain-of-Causal Reasoning, which parses and grounds intents through a functional causal chain that explicitly bridges abstract intents, functional requirements, and object suitability.
Specifically, our method first employs Intent Causal Parsing to decompose natural language intentions into functional requirements along a structured reasoning chain, as illustrated in the upper-right part of Fig. \ref{f1}.
By explicitly identifying and organizing latent needs, the module enables the model to more accurately capture the user’s intent.
For example, given the intent of natural language, ‘I want something to support my back to relieve pressure’, the model first identifies the core requirements: ‘support my back’ and ‘relieve pressure’. It then further infers the corresponding physical properties: the former requires the object to provide adequate support, while the latter implies that the object should be soft. Through this stepwise reasoning mechanism, the model transforms complex intentions into explicit functional requirements, forming a functional causal chain from abstract intent to functional requirements and then to object suitability.
This progressive functional causal reasoning allows the model to consistently understand complex intentions, reducing misunderstandings caused by semantic ambiguity.
Next, based on this reasoning chain, the parsed functional requirements are represented as intermediate nodes in a functional causal dependency graph, establishing explicit links from natural-language intent to object suitability and localization.
For example, the intent ‘support my back to relieve pressure’ is decomposed into functional nodes such as ‘supportiveness’ and ‘softness’, and functional dependency mappings are established between these functional nodes and candidate objects, as shown in the lower right part of Fig. \ref{f1}. 
This design not only mitigates the logical gaps present in traditional semantic-matching methods but also provides an interpretable functional causal reasoning process for 3D intent grounding, enhancing the model’s robustness and explainability in open-world scenarios.
Finally, we introduce Causal–Visual Feature Alignment, which encodes functional requirements as function-aware representations and aligns them with 3D point cloud features to enable bidirectional verification between functional reasoning and visual evidence.

Extensive experiments on 3D Intention Grounding and 3D Visual Grounding demonstrate that our method effectively enhances intent understanding and improves object localization.

To summarize, our contributions are as follows:

• We propose a new method, Chain-of-Causal Reasoning, which decomposes complex intents along a functional causal chain and builds an explicit intent–function–object dependency graph for interpretable and robust 3D object localization.

• We design a causal–visual feature alignment module that encodes functional requirements as functional causal representations and aligns them with 3D point cloud features, enabling bidirectional verification between functional causal reasoning and visual evidence.

• We validate the effectiveness of our method on 3D Intention Grounding and 3D Visual Grounding, demonstrating strong and competitive performance across multiple benchmarks.
\vspace{-10pt}
\section{Related Works}
\vspace{-5pt}

\subsection{Causal Graph Modeling}
\vspace{-5pt}
Causal modeling has been widely explored to improve model interpretability, robustness, and generalization. In visual tasks, existing studies introduce causal mechanisms for image classification~\cite{2DCausal}, object detection~\cite{2DCausal2}, point cloud classification~\cite{Caual}, and 3D novel class discovery~\cite{liyang}. In 3D visual grounding, MA$^2$TransVG~\cite{MAtrans} models causal effects among explicit textual attributes for fine-grained localization. However, these methods mainly focus on observable attributes, object categories, or point-level representations, while 3D intention grounding requires reasoning over latent functional needs behind abstract intents. Therefore, we focus on task-level functional causal reasoning by constructing an explicit intent--function--object causal chain. More discussions on functional requirements and object attributes are provided in Appendix~\ref{app:functional_affordance} and Appendix~\ref{Functional}.

\vspace{-8pt}
\subsection{3D Intention Grounding}
\vspace{-6pt}
3D Intention Grounding aims to localize target objects in 3D scenes according to natural-language human intentions. Different from conventional 3D visual grounding~\cite{Scanrefer,Referit3d,VPPNet,MBVG}, where queries usually contain explicit object categories, attributes, or spatial relations, intention grounding requires inferring implicit functional requirements from abstract and non-descriptive intents. Intent3D~\cite{intent3d} introduces this task and proposes IntentNet based on verb--object alignment. However, direct verb--object matching may introduce semantic logical gaps, since it does not explicitly explain why a candidate object satisfies the underlying functional needs of an intent. To address this limitation, we propose Chain-of-Causal Reasoning (CoCR), which decomposes complex intentions into functional requirements and builds a functional causal dependency graph for interpretable intent--function--object reasoning. Related 3D visual grounding and 3D-MLLM methods are discussed in Appendix~\ref{3D Visual Grounding}.
\vspace{-10pt}
\section{Method}
\begin{figure*}[!t]
  \centering
  \includegraphics[width=1.0\columnwidth]{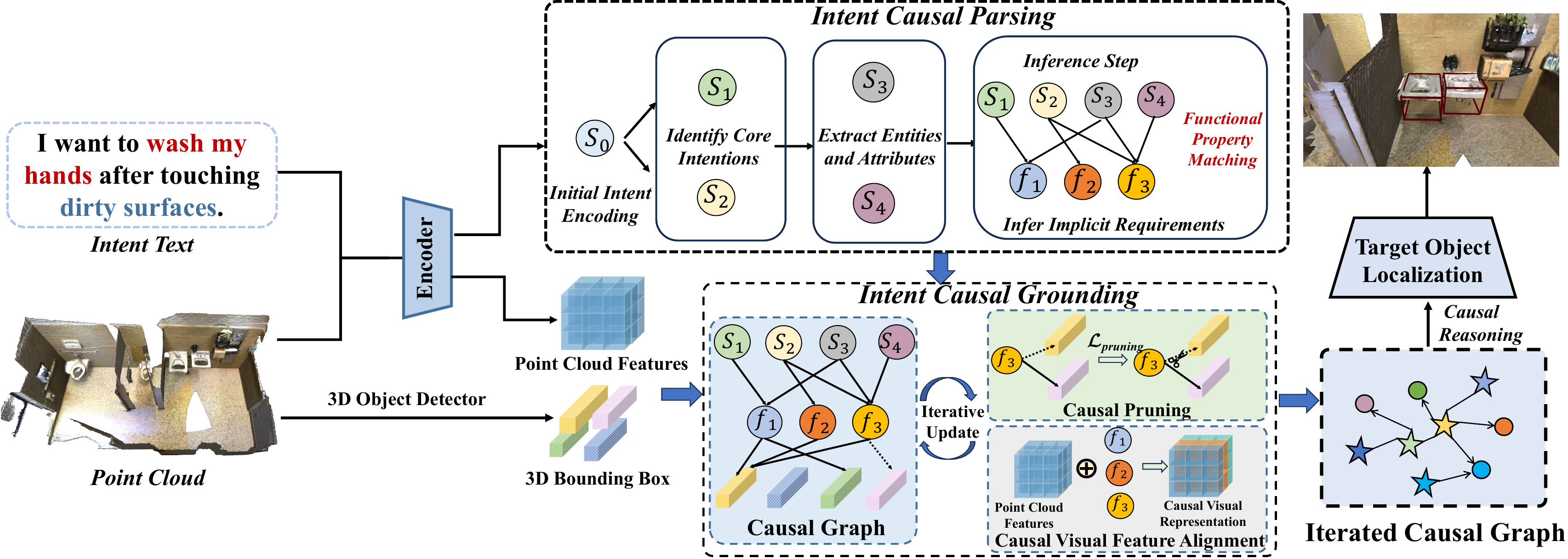}
  \vspace{-5pt}
  \caption{\textbf{The overall architecture}. 
    First, point cloud data and intent text are encoded by the model. The intent text is processed by the Intent Causal Parsing module for step-wise decomposition, transforming the natural language intent into functional requirements. Subsequently, Intent Causal Grounding is performed through the construction of a causal graph, defining the causal relationships between functional requirements and object attributes. The graph is iteratively updated via causal–visual feature alignment and causal pruning. Finally, causal reasoning is conducted based on the updated graph to achieve object selection and localization.}
    \label{F2}
     \vspace{-10pt}
\end{figure*}

\vspace{-8pt}
The overall framework of our Chain-of-Causal Reasoning (CoCR) method is illustrated in Fig. \ref{F2}. CoCR follows a functional causal reasoning process that explicitly bridges abstract intents, functional requirements, and candidate objects. The framework consists of three key modules: (i) Intent Causal Parsing, which decomposes complex intentions into step-wise functional requirements; (ii) Intent Causal Grounding, which establishes explicit functional dependency links between functional requirements and candidate objects; and (iii) Causal--Visual Feature Alignment, which iteratively updates the functional dependency graph and enables robust cross-modal reasoning. In CoCR, causal refers to task-level functional causal dependencies rather than formal causal identification under structural causal models; detailed discussion is provided in Appendix~\ref{app:causal_interpretation}.

\vspace{-10pt}
\subsection{Intent Causal Parsing}
\vspace{-5pt}
To extract actionable functional requirements from abstract intentions, we design the Intent Causal Parsing (ICP) module, which decomposes a natural-language intent into a step-wise functional causal chain from the original intent to latent needs and then to concrete functional requirements.

Given a natural-language intent $Q$, such as ‘I want something that can support my back to relieve pressure’, we employ a fine-tuned T5-small model~\cite{t5} with prompt templates to generate a query-conditioned reasoning sequence $\mathcal{S}$:
\begin{eqnarray}
\mathcal{S} = \{s_1, s_2, \dots, s_T\},
\end{eqnarray}
where $T$ denotes the number of reasoning steps and each $s_t$ represents an intermediate functional reasoning step generated conditioned on $Q$. For example, the intent ‘support my back and relieve pressure’ can be decomposed into core needs such as ‘back support’ and ‘pressure relief’, which are further associated with functional requirements such as supportiveness, softness, and comfort. More details about the prompt design and T5-small fine-tuning are provided in Appendix~\ref{mICP}.

To align the reasoning sequence with a unified functional space, each reasoning step $s_t$ is encoded into a vector $h_t \in \mathbb{R}^{d_s}$, where $d_s$ is set to 512. We then compare $h_t$ with each functional requirement node $f_j \in F$, where $F=\{f_1,\dots,f_K\}$ denotes the global functional requirement set. In our implementation, $F$ is constructed from the functional annotations and intent descriptions in the training split, followed by synonym merging and manual verification. The final vocabulary contains $K=40$ functional requirement nodes. The construction details of $F$ are provided in Appendix~\ref{mICP}. The cosine similarity is computed as:
\begin{eqnarray}
\cos(h_t, f_j) = \frac{h_t \cdot f_j}{\| h_t \| \| f_j \|},
\end{eqnarray}
where $\cdot$ denotes the dot product and $\|\cdot\|$ denotes the $L_2$ norm.
The similarity scores are normalized over all functional requirement nodes to obtain the step-level functional distribution:
\begin{eqnarray}
P(f_j \mid s_t) =
\frac{\exp(\cos(h_t, f_j))}
{\sum_{k=1}^{K} \exp(\cos(h_t, f_k))},
\end{eqnarray}
where $\exp(\cdot)$ denotes the exponential function, $K$ is the number of functional requirement nodes in $F$, and $P(f_j \mid s_t)$ represents the probability that the reasoning step $s_t$ corresponds to the functional requirement $f_j$.
Since all reasoning steps in $\mathcal{S}$ are generated conditioned on the original intent $Q$, each $P(f_j \mid s_t)$ can be regarded as a query-conditioned estimate of how strongly the $t$-th reasoning step requires the functional node $f_j$. We aggregate these step-level distributions to obtain the intent-level functional requirement distribution:
\begin{eqnarray}
P(f_j \mid Q) =
\frac{1}{T}\sum_{t=1}^{T} P(f_j \mid s_t),
\end{eqnarray}
where $T$ denotes the number of reasoning steps, and $P(f_j \mid Q)$ indicates the overall likelihood that the intent $Q$ requires the functional requirement $f_j$. This distribution serves as the bridge from abstract intent parsing to subsequent functional causal grounding. The parsing and grounding accuracy of the generated functional requirements is further evaluated in Appendix~\ref{Intent Parsing}.

\begin{algorithm}[tb]
   \caption{Chain-of-Causal Reasoning (CoCR) for 3D Intent Grounding}
   \label{alg:cocr_final}
\begin{algorithmic}[1]
   \STATE {\bfseries Input:} Point cloud $\mathcal{X}$, natural intent $Q$, candidate object set $\mathcal{O}$.
   \STATE {\bfseries Output:} Matched object $o^*$.
   
   \STATE {\bfseries Step 1: Intent Causal Parsing}
   \STATE Generate reasoning sequence $\mathcal{S}$ from intent $Q$ via T5-small using Eq. (1).
   \STATE Calculate step-level functional distributions $P(f_j|s_t)$ over the functional set $F$ using Eq. (2) and Eq. (3).
   \STATE Aggregate step-level distributions to obtain the intent-level functional distribution $P(f_j|Q)$ using Eq. (4).

   \STATE {\bfseries Step 2: Intent Causal Grounding}
   \STATE Construct a functional dependency graph $\mathcal{G}(\mathcal{V}, \mathcal{E})$ over $Q$, $F$, and $\mathcal{O}$.
   \STATE Initialize object--function dependencies $P(o_m|f_j)$ and compute $P(o_m|Q)$ using Eq. (5).
   \STATE Prune weak object--function edges where $w_{mj}<\theta$ using Eq. (6).
   \STATE Calculate the initial structural score $g_m=P(o_m|Q)$ using Eq. (7).

   \STATE {\bfseries Step 3: Causal--Visual Feature Alignment}
   \STATE Extract initial scene features $V$ from $\mathcal{X}$ via PointNet++.
   \STATE Generate function-aware representations $C$ weighted by $P(f_j|Q)$ using Eq. (9).
   \FOR{$n = 1$ \textbf{to} $N$}
        \STATE Perform visual--functional fusion: $\tilde{V} = \text{CrossAttn}(V, C)$ using Eq. (10).
        \STATE Compute dependency weights $a_{mj}$ for object--function pairs using Eq. (11).
        \STATE Update weights based on dependency-weighted visual--functional consistency using Eq. (12).
   \ENDFOR

   \STATE {\bfseries Step 4: Final Object Selection}
   \FOR{each candidate object $o_m \in \mathcal{O}$}
       \STATE Recompute $g_m$ using the updated object--function dependencies.
   \ENDFOR
   \STATE Select the optimal object using Eq. (8): $o^* = \arg\max_{m} g_m$.
    
   \STATE {\bfseries Return} $o^*$
\end{algorithmic}
\end{algorithm}

\subsection{Intent Causal Grounding}

After obtaining the functional requirement distribution $P(f_j \mid Q)$ from Intent Causal Parsing, we construct a functional causal dependency graph to perform Intent Causal Grounding. Explicitly modeling the functional dependencies among intents, functional requirement nodes, and candidate objects enables the model to map functional requirements to object suitability, mitigating the gaps caused by direct semantic matching.
Specifically, we map the functional requirement set $F = \{ f_1, f_2, \dots, f_K \}$ to intermediate nodes in the graph and the candidate object set $\mathcal{O} = \{ o_1, o_2, \dots, o_M \}$ to candidate terminal nodes, where $M$ denotes the total number of candidate objects. The graph $\mathcal{G} = (\mathcal{V}, \mathcal{E})$ consists of a node set $\mathcal{V} = \{ Q, F, \mathcal{O} \}$ and an edge set $\mathcal{E}$, where edges represent task-level functional dependency relationships. The intent node $Q$ is mapped to functional requirement nodes $f_j$, and the functional requirement nodes are further mapped to candidate object nodes $o_m$. The following conditional probability represents the structured dependency mapping:
\begin{eqnarray}
P(o_m \mid Q) = \sum_{j=1}^{K} P(o_m \mid f_j) \cdot P(f_j \mid Q),
\end{eqnarray}
where $P(f_j \mid Q)$ denotes the functional relevance of $f_j$ to the intent, and $P(o_m \mid f_j)$ represents the compatibility between candidate object $o_m$ and requirement $f_j$, initialized by a softmax over their text-embedding cosine similarities using predicted object category names. Since functional requirements are not mutually exclusive, Eq.~(5) performs a function-mediated weighted compatibility aggregation rather than applying the law of total probability.

To remove redundant functional links, we introduce a dependency pruning constraint~\cite{CR}. Let $w_{mj}$ denote the dependency weight between candidate object $o_m$ and functional requirement $f_j$, initialized by $P(o_m \mid f_j)$. We penalize weak dependencies whose weights are below a threshold $\theta$:
\begin{eqnarray}
\mathcal{L}_{pruning}(\theta) =
\sum_{m=1}^{M}\sum_{j=1}^{K}
\mathbb{I}(w_{mj}<\theta)\cdot w_{mj}^{2},
\end{eqnarray}
where $\mathbb{I}(\cdot)$ is the indicator function. During inference, edges with $w_{mj}<\theta$ are masked out to suppress unreliable object--function paths. This pruning strategy reduces redundant dependencies and improves graph reliability. We empirically set $\theta=0.2$ according to the ablation study.

Finally, based on the pruned functional dependency graph $\mathcal{G}$, we compute the structural matching score for each candidate object as:
\begin{eqnarray}
g_m = P(o_m \mid Q),
\end{eqnarray}
where $P(o_m \mid Q)$ aggregates the function-mediated dependencies from the intent $Q$ to candidate object $o_m$. The final matched object is selected by:
\begin{eqnarray}
o^*=\arg\max_m g_m.
\end{eqnarray}
By explicitly constructing and pruning the functional dependency graph, the model obtains initial function-mediated scores for candidate objects. These scores are subsequently refined through Causal--Visual Feature Alignment in Sec.~3.3. Eq.~(8) specifies the Top-1 selection rule based on the final candidate scores, whereas AP evaluates the ranked candidate predictions.

\subsection{Causal-Visual Feature Alignment}

After obtaining the functional dependency graph $\mathcal{G}$, we further integrate function-aware representations with the geometric--semantic features of the 3D scene point cloud to achieve collaborative reasoning between functional inference and visual perception.
Specifically, given the input 3D scene point cloud $\mathcal{X}$, we first employ a 3D point cloud encoder (e.g., PointNet++) to extract the scene representation $V \in \mathbb{R}^{M \times d}$, where $M$ denotes the number of candidate objects in the point cloud and $d$ is the feature dimension. Meanwhile, based on the functional requirements extracted in Intent Causal Parsing, represented by the probability distribution $P(f_j \mid Q)$, we obtain a set of function-aware representations $C \in \mathbb{R}^{K \times d}$, where $K$ corresponds to the number of functional requirements. These representations are computed as:
\begin{eqnarray}
C_j = P(f_j \mid Q) \cdot \text{Embed}(f_j), \quad j = 1, 2, \dots, K,
\end{eqnarray}
where $\text{Embed}(f_j)$ denotes the embedding of the $j$-th functional requirement $f_j$ in a $d$-dimensional vector space, and $P(f_j \mid Q)$ indicates its relevance to the input intent. The resulting set of function-aware representations $C$ is then used for fusion with 3D point cloud features.

To align functional requirements with candidate objects, we adopt a cross-attention mechanism:
\begin{eqnarray}
\tilde{V} = \text{CrossAttn}(V, C), 
\end{eqnarray}
where the visual features $V$ are used as queries, and the function-aware representations $C$ serve as keys and values to inject intent-related functional cues into candidate object features, resulting in fused visual-functional representations $\tilde{V}$.

On this basis, we introduce a dependency-weighted consistency constraint to enable mutual verification between functional reasoning and visual evidence. Specifically, we first compute the dependency weight between candidate object $o_m$ and functional requirement $f_j$ as:
\begin{eqnarray}
a_{mj} =
\frac{P(f_j \mid Q) P(o_m \mid f_j)}
{\sum_{m=1}^{M}\sum_{j=1}^{K} P(f_j \mid Q) P(o_m \mid f_j)}.
\end{eqnarray}
Then, the consistency loss is defined as:
\begin{eqnarray}
\mathcal{L}_{\text{cons}} =
\sum_{m=1}^{M}\sum_{j=1}^{K}
a_{mj}
\left(1-\cos(\tilde{v}_m, C_j)\right),
\end{eqnarray}
where $\tilde{v}_m$ denotes the updated visual feature of candidate object $o_m$, and $C_j$ denotes the function-aware representation of the $j$-th functional requirement. Different from uniformly pulling all candidate features toward all functional representations, the dependency weight $a_{mj}$ enables selective consistency learning. Thus, only functionally relevant object--requirement pairs are encouraged to be close, while irrelevant candidates receive negligible weights. This design preserves object discriminability while improving the consistency between functional reasoning and visual evidence.

Unlike directly computing matching scores, the fused representations are further utilized to iteratively update the functional dependency graph $\mathcal{G}$, dynamically adjusting node embeddings and edge weights based on the consistency between visual and functional information. Through this iterative updating process, the dependency graph evolves alongside the fused representations, reinforcing reliable associations between functional requirements and candidate objects. Specifically, after $N$ iterations, the final candidate scores are recomputed using the updated function--object dependency weights, allowing the refined visual evidence to directly influence the final grounding predictions. More reasoning details, including the specific iterative mechanism (set to $N=2$), can be found in Appendix \ref{Causal-Graph-Based Reasoning} and \ref{supp:iterative_details}.

\section{Experiments}
 \vspace{-5pt}

In the experiments, for the 3D Intention Grounding (3D-IG) task, we follow the settings of Intent3D \cite{intent3d} to evaluate the model’s cross-modal reasoning ability between natural language intention understanding and 3D object grounding. We conducted comparative experiments on both the validation and test sets and performed fair comparisons with multiple baseline models. Additionally, we conducted further evaluations on 3D Visual Grounding task to comprehensively assess the proposed method’s effectiveness in terms of accuracy and robustness.

     
 \vspace{-5pt}
\subsection{Experimental setup}
 \vspace{-5pt}
\textbf{Dataset.} In our experiments, for the 3D Intention Grounding (3D-IG) task, we adopt the official evaluation setting of the Intent3D dataset \cite{intent3d}. Intent3D is primarily derived from the ScanNet \cite{scannet} dataset and contains 44,990 intention texts across 1,042 ScanNet scenes. The dataset covers large-scale indoor 3D environments paired with natural language intention descriptions, providing an effective benchmark for assessing a model’s cross-modal reasoning ability between intention understanding and 3D object grounding. Additionally, we validate the effectiveness of our method on the Nr3D, Sr3D, and ScanRefer datasets for the 3D Visual Grounding task, with more detailed descriptions of these datasets provided in Appendix~\ref{data}.

\textbf{Evaluation Metric.} 
Consistent with baseline methods, we evaluate our model using Top-1 Accuracy and Average Precision (AP). Top-1 Accuracy measures the correctness of the model’s highest-confidence prediction, while AP assesses detection performance across varying confidence thresholds. In experiments, we compute IoU between predicted and ground-truth boxes, and count a prediction as correct if IoU exceeds 0.25 or 0.5.

 \vspace{-5pt}
\subsection{Implementation Details}
 \vspace{-5pt}
To ensure a fair comparison, we adopt the same point cloud feature extractor, PointNet++ \cite{pointnet++}, and the text feature extractor, RoBERTa \cite{roberta}, as IntentNet. The model is trained for a total of 120 epochs on four NVIDIA 3090 GPUs with 24 GB of memory, using a batch size of 16, and the overall training time is approximately 40 hours. In terms of input settings, the number of point cloud tokens is fixed at 1024, and the maximum length of text tokens is set to 256. During training, our method parses intention causal chains via Intent Causal Parsing and builds a causal graph through Intent Causal Grounding to align functional requirements with object features, enabling causal reasoning. More implementation details are provided in the Appendix \ref{Imple}.

\begin{table*}[h]
  \centering
  \vspace{-10pt}
  \caption{3D Intention Grounding results on Intent3D validation and test sets. $^0$ indicates zero-shot results. The best results are in \textbf{bold}, and the second-best results are \underline{underlined}.}
   \label{Table1}
  \resizebox{\textwidth}{!}{
    \begin{tabular}{l c cccc cccc}
      \toprule
      \toprule
      \multirow{2}{*}{\textbf{Method}} 
      & \multirow{2}{*}{\textbf{Detector}} 
      & \multicolumn{4}{c}{\textbf{Val Set}} 
      & \multicolumn{4}{c}{\textbf{Test Set}} \\
      \cmidrule(lr){3-6} \cmidrule(lr){7-10}
      & 
      & \textbf{Top1@0.25} 
      & \textbf{Top1@0.5} 
      & \textbf{AP@0.25} 
      & \textbf{AP@0.5}
      & \textbf{Top1@0.25} 
      & \textbf{Top1@0.5} 
      & \textbf{AP@0.25} 
      & \textbf{AP@0.5} \\
      \midrule
      BUTD-DETR~\cite{BUTD} 
      & GroupFree~\cite{GroupFree} 
      & 47.12 & 24.56 & 31.05 & 13.05
      & 47.86 & 25.74 & 31.41 & 13.46 \\

      EDA~\cite{EDA} 
      & GroupFree~\cite{GroupFree} 
      & 43.11 & 18.91 & 14.02 & 5.00
      & 44.00 & 19.62 & 14.56 & 5.18 \\

      3D-ViSTA~\cite{3DVisTA} 
      & Mask3D~\cite{Mask3D} 
      & 42.76 & 30.37 & 36.10 & 19.93
      & 43.88 & 31.44 & 37.29 & 22.00 \\

      Chat-3D-v2$^0$~\cite{Chat-3d} 
      & PointGroup~\cite{pointgroup} 
      & 5.86 & 5.24 & 0.15 & 0.13
      & 5.63 & 4.93 & 0.14 & 0.11 \\

      Chat-3D-v2~\cite{Chat-3d} 
      & PointGroup~\cite{pointgroup} 
      & 36.71 & 32.78 & 3.23 & 2.58
      & 33.46 & 29.32 & 2.67 & 2.10 \\

      IntentNet~\cite{intent3d} 
      & GroupFree~\cite{GroupFree} 
      & \underline{58.34} & \underline{40.83} & \underline{41.90} & \underline{25.36}
      & \underline{58.92} & \underline{42.28} & \underline{44.01} & \underline{27.60} \\
      \midrule
      Ours 
      & GroupFree~\cite{GroupFree} 
      & \textbf{60.71} & \textbf{43.24} & \textbf{42.65} & \textbf{27.33}
      & \textbf{61.37} & \textbf{45.91} & \textbf{45.73} & \textbf{30.28} \\
      \bottomrule
      \bottomrule
    \end{tabular}
    
  }
  \vspace{-5pt}
\end{table*}

 \vspace{-5pt}
\subsection{Comparison with the State of the Art}
 \vspace{-5pt}
\textbf{Validation on 3D Intention Grounding.}
Table \ref{Table1} presents a quantitative comparison of our method on the Intent3D validation and test sets. Following the Intent3D evaluation protocol, we compare three categories of approaches: expert models (e.g., BUTD-DETR \cite{BUTD} and EDA \cite{EDA}), foundation models (e.g., 3D-VisTA \cite{3DVisTA}), and LLM-based models (e.g., Chat-3D-v2 \cite{Chat-3d}), as well as the IntentNet method. As shown in Table \ref{Table1}, our approach achieves the best performance in both Top-1 Accuracy and Average Precision. For instance, at an IoU threshold of 0.5, our method improves Top-1 Accuracy on the validation set by 6\% compared to IntentNet.
Our method has three main advantages: Intent Causal Parsing decomposes complex intents into causally linked functional requirements; Intent Causal Grounding precisely maps these requirements to object attributes while removing redundant paths; Causal-Visual Feature Alignment strengthens causal–visual interaction via cross-attention and bidirectional consistency, reducing cross-modal misalignment.
We further compare our approach with a two-stage method that directly leverages a large language model for intent analysis and object localization; details are provided in Appendix \ref{llm}.

\begin{figure*}[!t]
  \centering
  \includegraphics[width=1.0\columnwidth]{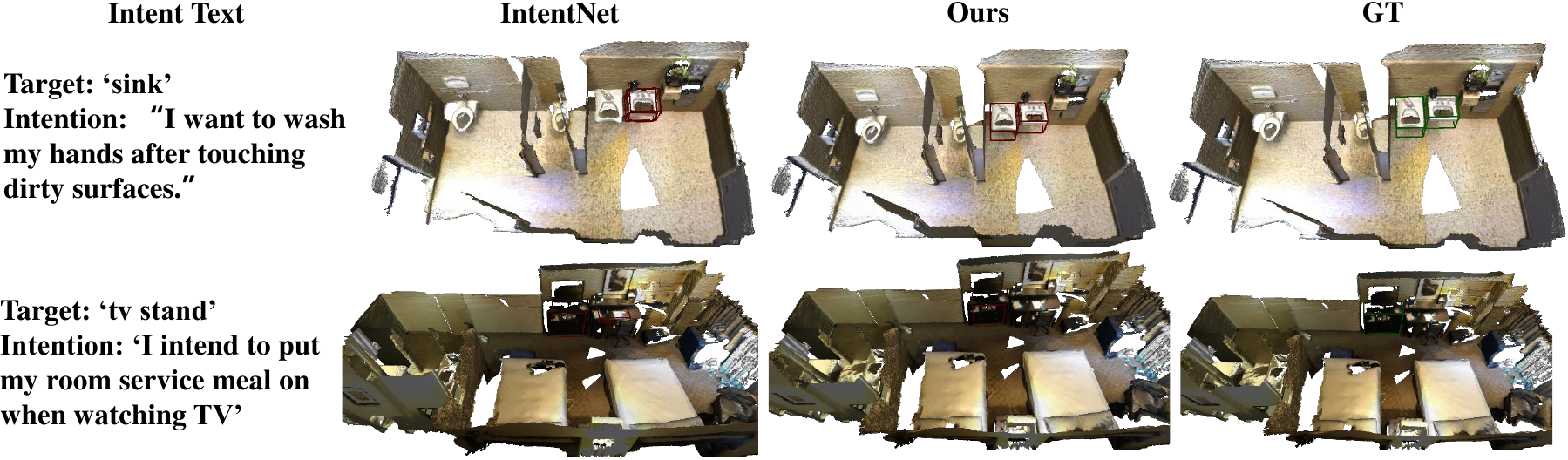}
  \caption{\textbf{Visualization comparison between IntentNet and our method on the Intent3D dataset.} As shown in the figure, our method on the Intent3D dataset demonstrates superior accuracy. }
    \label{vis}
    \vspace{-15pt}
\end{figure*}



\textbf{Visualization Analysis.}  Figure \ref{vis} presents a visual comparison between our method and IntentNet. The results clearly demonstrate that our approach achieves higher accuracy in both intent parsing and target object localization. For instance, in the first row of Figure \ref{vis}, our method successfully identifies and localizes all the sinks in the scene. In contrast, IntentNet only detects one, highlighting our model’s superior capability in comprehensive intent understanding. Furthermore, as shown in the second row, our method precisely localizes the target object, the TV stand, while effectively excluding other functionally similar objects, such as the table. This advantage stems from our accurate intent parsing and functional matching, which together ensure precise final localization. Further visual and failure case analyses are provided in Appendix \ref{Avis} and \ref{FCA}.

\begin{wraptable}{r}{0.52\textwidth}
\centering
 \vspace{-20pt}
\caption{Performance Comparison on 3D Visual Grounding for Nr3D and Sr3D Datasets.}
 \fontsize{22}{23}\selectfont
\resizebox{1.0\linewidth}{!}{
\begin{tabular}{lcccc}
\toprule
\toprule
    \multirow{2}{*}{\textbf{Methods}} & \multicolumn{2}{c}{\textbf{Nr3D}} & \multicolumn{2}{c}{\textbf{Sr3D}} \\
    \cmidrule(lr){2-3} \cmidrule(lr){4-5}
    & \textbf{Overall} & \textbf{Hard} & \textbf{Overall} & \textbf{Hard} \\
    \midrule
    ReferIt3D \cite{Referit3d} & 35.6 & 27.9 & 40.8 & 31.5 \\
    TGNN \cite{TGNN} & 37.3 & 30.6 & 45.0 & 36.9 \\
    3D-SPS \cite{3dsps}& 51.5 & 45.1 & 62.6 & 65.4 \\
    EDA \cite{EDA} & 52.1 & 46.1 & 68.1 & 62.9 \\
    MVT \cite{MVT} & 55.4 & 49.1 & 64.5 & 58.8 \\
    G$^3$-LQ \cite{G3LQ} & 58.4 & 50.7 & 73.1 & 66.3 \\
    MiKASA \cite{mikasa}& 64.4 & 59.4 & 75.2 & 67.3 \\
    GPS \cite{GPS}& 64.9 & 57.8 & 77.5 & 71.6 \\
    MA$^2$TransVG \cite{MAtrans} & 65.2 & 57.6 & 73.9 & 69.3 \\
    PQ3D \cite{PQ3Q} & 66.7 & 58.7 & 79.7 & 72.8 \\
    \midrule
    Ours & \textbf{67.1} & \textbf{59.8} & \textbf{81.6} & \textbf{73.7} \\
    \bottomrule
    \bottomrule
\end{tabular}

}
\vspace{-10pt}
\label{vg}
\end{wraptable}
\textbf{Validation on 3D Visual Grounding.}
To further evaluate the generality of our method, we conduct experiments on the 3D Visual Grounding task. In this setting, Intent Causal Parsing mainly performs step-wise text decomposition to extract informative semantic cues from descriptive expressions, while Intent Causal Grounding and Causal--Visual Feature Alignment construct functional dependency relations and align them with 3D visual features. As shown in Table~\ref{vg}, our method achieves the best results in both overall accuracy and challenging cases. The improvement indicates that explicit functional dependency modeling can enhance cross-modal alignment and help the model distinguish visually similar objects in cluttered scenes. Additional results on ScanRefer~\cite{Scanrefer} and component-wise ablations on 3D Visual Grounding are provided in Appendix~\ref{ScanRefer} and Appendix~\ref{Ab2}, respectively.

\vspace{-10pt}
\subsection{Ablation Study}
\vspace{-5pt}
\textbf{Component Analysis.}
We conduct ablation experiments on the Intent3D test set to evaluate the contributions of the three core modules: Intent Causal Parsing (ICP), Intent Causal Grounding (ICG), and Causal--Visual Feature Alignment (CVFA), as summarized in Table~\ref{ab}. The baseline in the first row directly performs intent--object matching. Adding ICP brings a clear 
\begin{wraptable}{r}{0.52\textwidth}
  \centering
\vspace{-7pt}
  \caption{Ablation analysis of our proposed model. ICP stands for Intent Causal Parsing, ICG represents Intent Causal Grounding, and CVFA denotes Causal-Visual Feature Alignment.}
  \resizebox{1.0\linewidth}{!}{
    \begin{tabular}{ccc|cccc}
      \toprule
      \toprule
      \multicolumn{3}{c|}{\textbf{Methods}} & \multicolumn{4}{c}{\textbf{Metrics}} \\
      \cmidrule(r){1-3}\cmidrule(l){4-7}
      \textbf{ICP} & \textbf{ICG} & \textbf{CVFA} &
      \textbf{Top1-Acc@0.25} & \textbf{Top1-Acc@0.5} & \textbf{AP@0.25} & \textbf{AP@0.5} \\
      \midrule
       & & &  55.32 & 38.70 & 40.72 & 24.40\\
       $\checkmark$ & & &57.82 & 40.31 &42.85&25.70 \\
       & $\checkmark$ & & 58.68 & 41.07 & 43.76 & 25.37 \\
        & $\checkmark$ & $\checkmark$ &  59.28 & 42.46 & 43.71 &27.01 \\
      $\checkmark$ & $\checkmark$ & & 60.99 & 43.67 & 44.25 & 28.38  \\
    
      $\checkmark$ & $\checkmark$ & $\checkmark$ & \textbf{61.37} & \textbf{45.91} & \textbf{45.73} & \textbf{30.28} \\
      \bottomrule
      \bottomrule
    \end{tabular}
    
  }
\vspace{-5pt}
  \label{ab}
\end{wraptable}
improvement, showing that decomposing abstract intentions into hierarchical functional requirements helps the model better capture the underlying user needs even without explicit graph reasoning.
When ICP is removed, ICG relies only on core verbs and objects extracted by the text encoder to construct the functional dependency graph. This setting slightly improves over the baseline, but the gain is limited because direct verb--object cues still introduce semantic logical gaps.
Further adding CVFA improves the performance by aligning functional dependency representations with 3D visual features, validating the benefit of causal--visual interaction. Combining ICP and ICG yields larger gains than using either module alone, indicating that step-wise intent parsing and explicit functional dependency modeling are complementary. Finally, the full model achieves the best performance after introducing CVFA, demonstrating that the three modules jointly contribute to more accurate and robust 3D intention grounding.

\begin{wraptable}{r}{0.52\textwidth}
\centering
\vspace{-20pt}
\caption{Efficiency comparison in training time, inference latency, memory, and parameters.}
\label{efficiency}
\resizebox{\linewidth}{!}{
\begin{tabular}{lcccc}
\toprule
\toprule
\multirow{2}{*}{\textbf{Method}} &
\textbf{Training Time} &
\textbf{Inference} &
\textbf{Memory} &
\textbf{Params} \\
& \textbf{(hours)} & \textbf{(ms)} & \textbf{(GB)} & \textbf{(M)} \\
\midrule
IntentNet~\cite{intent3d} & 32 & 27.1 & 7.1 & 148.6 \\
Baseline & 26 & 25.9 & 6.8 & 147.3 \\
Baseline + ICP & 34 & 28.4 & 7.3 & 207.8 \\
Baseline + ICP + ICG & 38 & 29.1 & 7.6 & 208.6 \\
Full Model (Ours) & 40 & 32.7 & 7.6 & 209.5 \\
\bottomrule
\bottomrule
\end{tabular}
}
\vspace{-10pt}
\end{wraptable}

\textbf{Efficiency Analysis.}
Table~\ref{efficiency} reports the efficiency comparison of different variants. Training Time is measured on four 24GB RTX 3090 GPUs with batch size 16, while Inference denotes the average latency per sample on a single 24GB RTX 3090 GPU. Compared with the baseline, ICP introduces additional parameters and latency due to the T5-small parser for intent decomposition. ICG only slightly increases the overhead, since it mainly performs lightweight functional dependency construction over a fixed functional vocabulary. CVFA further introduces visual--functional alignment, leading to a moderate increase in inference latency. Overall, the full CoCR model maintains acceptable memory usage and single-GPU deployability, while providing stronger structured reasoning for 3D intention grounding.

 \begin{wrapfigure}{r}{0.4\columnwidth}
  \centering
    \vspace{-15pt}
  \includegraphics[width=0.4\columnwidth]{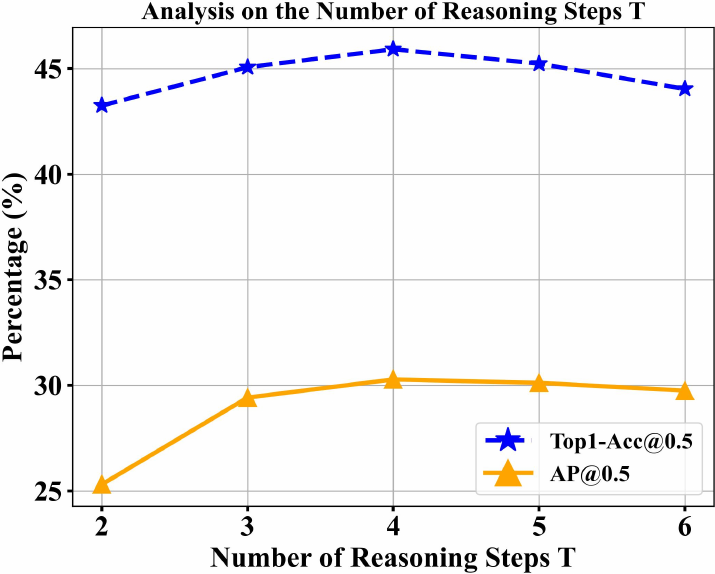}
  \vspace{-20pt}
  \caption{Analysis of the reasoning steps number $T$ in Intent Causal Parsing.}
    \label{hyper}
\vspace{-15pt}
\end{wrapfigure}
\textbf{Analysis of the Number of Reasoning Steps $T$ in Intent Causal Parsing.} In Intent Causal Parsing, the number of reasoning steps is a key hyperparameter. Too few steps result in incomplete parsing, while too many add complexity, increase computational cost, and risk error propagation. As shown in Figure \ref{hyper}, the model achieves its best performance on the Intent3D test set with four steps, reaching the highest Top1-Acc@0.5 and AP@0.5. 
The model effectively extracts core functional requirements and matches them with target objects. Increasing the reasoning steps beyond four introduces redundant information, complicates causal graph construction, and may degrade performance.
These findings emphasize the need to optimize the number of steps for balancing parsing depth and efficiency. The effect of the causal graph update iterations $N$ in CVFA is further analyzed in Appendix~\ref{supp:iterative_details} and \ref{Causal Graph Update Iterations}.


\begin{wraptable}{r}{0.52\textwidth}
  \centering
  \vspace{-20pt}
  \caption{Structural perturbation analysis on the Intent3D test set.}
  \resizebox{1.0\linewidth}{!}{
    \begin{tabular}{lcccc}
      \toprule
      \toprule
      \textbf{Perturbation} & \textbf{Top1@0.25} & \textbf{Top1@0.5} & \textbf{AP@0.25} & \textbf{AP@0.5} \\
      \midrule
      Full Model & \textbf{61.37} & \textbf{45.91} & \textbf{45.73} & \textbf{30.28} \\
      Top-contribution removal & 52.51 & 35.85 & 34.13 & 21.45 \\
      Random removal & 58.45 & 42.39 & 42.27 & 27.57 \\
      Edge shuffling & 54.56 & 38.28 & 39.61 & 24.32 \\
      \bottomrule
      \bottomrule
    \end{tabular}
  }
  \label{tab:structural_perturbation}
\end{wraptable}

\textbf{Structural Perturbation Analysis.}
To evaluate the contribution of functional dependencies to final predictions, we conduct inference-time structural perturbations, including top-contribution and random node/edge removal under the same budget, as well as edge shuffling with the edge count preserved. As shown in Table~\ref{tab:structural_perturbation}, removing high-contribution nodes and edges reduces Top1@0.5 from 45.91\% to 35.85\%, compared with 42.39\% under random removal. Edge shuffling further decreases performance to 38.28\%, demonstrating that CoCR relies on specific functional intermediates and dependency structures rather than arbitrary graph connections.

\begin{wraptable}{r}{0.52\textwidth}
  \centering
\vspace{-7pt}
  \caption{Impact of causal pruning threshold $\theta$.}
  
  \resizebox{1.0\linewidth}{!}{
    \begin{tabular}{ccccccc}
      \toprule
      \toprule
      Pruning Threshold ($\theta$) & 0.1 & 0.2 & 0.3 & 0.4 & 0.5 \\
      \midrule
      AP@0.5 & 25.41 & \textbf{27.33} & \underline{25.82} & 22.35 & 21.92 \\
      Top1-Acc@0.5 & 41.75 & \textbf{43.24} & \underline{42.63} & 40.89 & 38.41 \\
      \bottomrule
      \bottomrule
    \end{tabular}
  }
  \label{pruning_ablation}
\vspace{-10pt}
\end{wraptable}
\textbf{Analysis of the Causal Pruning Threshold $\theta$.} To evaluate the effectiveness of our causal pruning strategy and find the optimal threshold, we conducted an ablation study on the pruning threshold $\theta$.
As shown in Table \ref{pruning_ablation}, both overly low and overly high thresholds have a negative impact on performance. When $\theta$ is too small (e.g., 0.1), the causal graph retains excessive redundant paths, and the introduced noise interferes with accurate reasoning. Conversely, an overly high threshold (e.g., 0.4, 0.5) over-prunes the graph, removing effective causal links that are crucial for matching key functional attributes. Our model achieved its peak performance (AP@0.5) at $\theta=0.2$. This indicates that this threshold strikes the best balance between removing irrelevant information and preserving necessary causal connections. This finding validates the importance of our pruning mechanism, as it optimizes the causal graph structure, thereby enhancing the model's accuracy and efficiency.




\section{Conclusion}

For the 3D Intention Grounding (3D-IG) task, we propose Chain-of-Causal Reasoning (CoCR), a causality-inspired functional dependency reasoning framework for intent-aware 3D object localization. Specifically, the method incrementally decomposes complex intentions into functional requirements along the causal chain, clarifying priorities and latent needs, thus forming a causal chain from abstract intentions to object attributes and improving the accuracy of intent understanding. Based on this, we build a causal graph linking functional requirements to object attributes and employ a causal-visual fusion mechanism for deep cross-modal alignment and bidirectional verification. Extensive experiments on 3D Intention Grounding and 3D Visual Grounding tasks demonstrate that our method effectively enhances intent understanding and improves object localization accuracy.

\paragraph{Acknowledgment.} This work was supported by the National Natural Science Foundation of China (Grant No. 62636008, 62376186, 62472333, and 61932009).

\bibliographystyle{IEEEtran}
\bibliography{main}
\newpage


\appendix
\section{Appendix}
\subsection{Related Works}
\subsubsection{Functional Affordance and Causality-inspired Reasoning}
\label{app:functional_affordance}

The 3D Intention Grounding task is closely related to functional affordance understanding, since the target object is determined by whether it can satisfy the user's intended function. The concept of affordance emphasizes the actionable properties or functional possibilities provided by an object to an agent~\cite{Affordances}. In visual understanding, affordance and function recognition have been widely studied to infer what actions or utilities an object can support beyond its visual appearance~\cite{hassanin2021visual}. Recent studies further explore affordance grounding with vision-language models~\cite{Affordancellm} and 3D object affordance grounding from language instructions, visual observations, and interactions~\cite{grounding3d}, showing the importance of connecting language, perception, and object functionality.
However, 3D Intention Grounding differs from conventional affordance understanding in its problem formulation. Most affordance-related methods start from observed objects, visual regions, or language instructions, and then predict possible actions, interaction regions, or object functions. In contrast, 3D Intention Grounding starts from an abstract human intent and requires the model to infer the latent functional requirements before selecting the target object. For example, the intent ‘I want something to support my back and relieve pressure’ does not directly specify an object category or an interaction region. Instead, the model must infer functional requirements such as supportiveness, softness, and comfort, and then determine which candidate object in the current 3D scene can satisfy these requirements.

Our CoCR framework can be viewed as a causality-inspired extension of functional affordance reasoning for 3D intention grounding. Specifically, functional requirements are introduced as intermediate causal variables between abstract intentions and candidate objects. The intent induces latent functional requirements, and these requirements further determine the suitability of candidate objects. This forms an explicit intent--function--object causal chain, which differs from direct intent--object similarity matching. By introducing functional requirements as mediators, CoCR provides an interpretable reasoning mechanism that explains why a candidate object is suitable for a given intention.
This formulation also clarifies the relation between our method and strict causal inference. We do not claim to estimate causal effects under structural causal models. Instead, we focus on task-level functional causal dependencies, where the causal meaning is reflected by functional necessity and directional reasoning. In this sense, CoCR bridges affordance-based object understanding and causality-inspired structured reasoning, making it more suitable for abstract 3D intention grounding where explicit visual descriptions are absent.
\subsubsection{3D Visual Grounding}
\label{3D Visual Grounding}
3D Visual Grounding aims to localize target objects in a 3D scene according to natural-language descriptions. Existing methods mainly focus on aligning linguistic expressions with object-level or point-level 3D representations. Early benchmarks such as ScanRefer~\cite{Scanrefer} and ReferIt3D~\cite{Referit3d} establish the basic setting of language-guided 3D object localization, where the input query usually contains explicit object categories, attributes, or spatial relations. Based on this setting, a series of methods have been proposed to improve cross-modal alignment. TGNN~\cite{TGNN} introduces text-guided graph neural networks to model object relations, while 3D-SPS~\cite{3dsps} performs single-stage grounding through referred point progressive selection. EDA~\cite{EDA} further decouples textual expressions and conducts dense alignment between language and 3D proposals. Recent works also enhance grounding performance through multi-view modeling, semantic-geometric alignment, and scene-aware representations, such as MVT~\cite{MVT}, G$^3$-LQ~\cite{G3LQ}, MiKASA~\cite{mikasa}, GPS~\cite{GPS}, and PQ3D~\cite{PQ3Q}. In addition, TSP3D~\cite{vg2} improves grounding efficiency by using text-guided sparse voxel pruning and completion.
Beyond single-object grounding, some studies extend 3D visual grounding to more complex scenarios. Multi3DRefer~\cite{M3DRef} focuses on grounding text descriptions to multiple 3D objects, while ConcreteNet~\cite{ConcreteNet} improves verbo-visual fusion for dense 3D visual grounding. D-LISA~\cite{D-LISA} introduces dynamic modules and language-informed spatial attention for multi-object grounding. These methods show that modeling object relations, spatial context, and language-guided interaction is important for complex 3D grounding. However, they still largely assume that the language query provides descriptive cues that can be directly aligned with visual objects.
Recently, 3D multimodal large language models have been introduced to grounded 3D scene understanding. Representative methods such as Chat-Scene~\cite{Chat-scene}, Chat-3D~\cite{Chat-3d}, and Robin3D~\cite{Robin3d} leverage the reasoning ability of large language models to connect 3D scene representations with natural-language instructions. These methods show strong potential for general 3D vision-language understanding, but they often require explicit object identifiers, object categories, or descriptive instructions to support reliable localization. In contrast, 3D Intention Grounding focuses on abstract and non-descriptive human intentions, where the target object is implied by latent functional requirements rather than directly mentioned in the query. Therefore, this task requires not only visual-language alignment but also functional reasoning from human intent to object suitability.

\subsubsection{Functional Requirements vs. Object Attributes} 
\label{Functional}

To clarify the specific challenge of 3D Intention Grounding, it is important to distinguish functional requirements from object attributes. Object attributes refer to explicit and observable properties of an object, such as color, shape, size, material, category, or spatial relation. These attributes are widely used in conventional 3D visual grounding, since the language query often directly describes the target object. For example, a query such as ‘the red chair near the table’ provides clear visual and spatial cues, allowing the model to localize the target through direct text--object alignment.
In contrast, functional requirements describe the expected capability or utility of an object, such as supportiveness, softness, washability, sittability, storability, or comfort. These requirements are usually not directly observable from the surface form of the text, but must be inferred from abstract human intentions. For example, the intent ‘I want something to support my back and relieve pressure’ does not explicitly mention a target category or visual attribute. Instead, it implies latent functional requirements such as supportiveness, softness, and comfort. Therefore, 3D Intention Grounding requires the model to infer what functions are needed and then determine which candidate objects can satisfy these functions in the current 3D scene.
This distinction differentiates our method from attribute-centric grounding methods. For instance, MA$^2$TransVG~\cite{MAtrans} models multi-attribute interactions and employs an Attribute Causal Analysis Module to quantify the causal effects of explicit visual attributes extracted from the text. This mechanism is effective for descriptive queries that contain rich attribute cues, such as color, shape, size, or spatial relation. However, abstract intention queries often lack such explicit descriptions, making attribute-level causal modeling insufficient for intention grounding. Similarly, methods such as VPPNet~\cite{VPPNet}, MBVG~\cite{MBVG}, and PQ3D~\cite{PQ3Q} improve visual-language grounding through stronger feature interaction or promptable queries, but their grounding process still mainly relies on explicit linguistic-visual correspondences.

Unlike these methods, our Chain-of-Causal Reasoning framework introduces functional requirements as intermediate causal variables between abstract intentions and candidate objects. Instead of directly matching an intent with object categories or visual attributes, CoCR first decomposes the intent into latent functional requirements and then constructs an intent--function--object reasoning chain. In this way, the model can explicitly explain why a candidate object is suitable for a given intention. For example, a pillow may be selected not because the query mentions ‘pillow’, but because it satisfies the inferred requirements of softness and pressure relief. This functional causal reasoning provides a more suitable mechanism for 3D Intention Grounding, especially when the input intent lacks explicit visual descriptions.

\subsection{More Method Details}

\subsubsection{Causal Interpretation of CoCR}
\label{app:causal_interpretation}

We clarify that the term ‘causal’ in CoCR refers to task-level functional causal dependencies in 3D intention grounding, rather than formal causal identification under structural causal models (SCMs), do-calculus, or intervention-based causal inference. In this work, causality is defined from the perspective of functional necessity: an abstract human intent gives rise to a set of latent functional requirements, and these functional requirements further determine whether a candidate object is suitable for satisfying the intent. Therefore, the causal relation modeled by CoCR is not a statistical treatment effect, but a directional functional dependency chain from intent to function and then to object suitability.

We do not formulate CoCR as a strict SCM for the following reasons. First, the goal of 3D intention grounding is not to estimate the causal effect of an intervention, but to infer which object best satisfies an abstract human intention in a 3D scene. A formal SCM usually requires explicitly defined endogenous variables, exogenous noise variables, structural equations, and intervention operators. However, in 3D-IG, the key variables, such as user intention, functional requirements, object affordance, scene context, and object suitability, are often latent, semantic, and task-dependent, making it difficult to specify universally valid structural equations. Second, existing 3D-IG datasets provide paired intention texts, 3D scenes, and target object annotations, but they do not contain interventional data, counterfactual annotations, or controlled manipulations of functional requirements. As a result, formal causal identification or do-calculus-based estimation is not supported by the available supervision. Third, the functional relation between an intent and an object is highly context-dependent. For example, an object that satisfies ‘support my back’ in one scene may be unsuitable in another scene due to its pose, accessibility, size, or surrounding context. This makes it more appropriate to model task-level functional dependencies than to assume a fixed global causal structure.

Specifically, given an intention query $Q$, CoCR first decomposes it into a set of intermediate functional requirements $F=\{f_1,\dots,f_K\}$. These functional requirements serve as causal mediators between the abstract intent and candidate objects. For example, the intent ‘I want something to support my back and relieve pressure’ does not directly indicate a specific object category. Instead, it implies functional causes such as supportiveness, softness, and comfort. These functional causes then determine why objects such as a chair, sofa, or pillow may be considered suitable, while other visually or semantically related objects may be rejected. In this sense, CoCR explicitly models the reasoning path $Q \rightarrow F \rightarrow O$, where $F$ explains the functional causes behind object selection.

The proposed functional causal dependency graph further operationalizes this idea by assigning directional dependency weights between the intent, functional requirements, and candidate objects. The edge from $Q$ to $f_j$ reflects how strongly the intent requires a certain function, while the edge from $f_j$ to $o_m$ reflects how well the candidate object satisfies that functional requirement. The final object score is therefore not obtained through direct intent--object matching, but through the aggregation of function-mediated causal dependencies. This design reduces the logical leap of direct semantic matching and provides an explicit intermediate reasoning structure.

Although CoCR does not claim to estimate causal effects in the strict SCM sense, its causal nature is reflected in three aspects. First, it introduces functional requirements as explicit mediating variables, which explain why an intent should lead to a particular object. Second, the dependency graph is directional: the intent induces functional requirements, and functional requirements determine object suitability, rather than treating all text and object features as undirected correlations. Third, the graph supports intervention-style analysis at the functional-node level. For instance, suppressing a key functional node such as ‘softness’ would reduce the suitability score of objects whose relevance mainly comes from pressure relief, while increasing the weight of ‘supportiveness’ would strengthen the scores of objects that provide stable body support. Such node-level perturbations reveal how functional causes affect the final grounding decision.

Therefore, the ‘causal’ in Chain-of-Causal Reasoning should be understood as task-level functional causal reasoning: CoCR makes the latent functional causes behind object suitability explicit, organizes them into a directional intent--function--object chain, and verifies them with 3D visual evidence through causal--visual feature alignment. This formulation is weaker than formal causal inference, but stronger than direct similarity-based matching, because it exposes the functional mechanisms that connect abstract intentions to concrete 3D objects.

\subsubsection{More Details about Intent Causal Parsing}
\label{mICP}

\begin{figure*}[h]
  \centering
  \includegraphics[width=1.0\columnwidth]{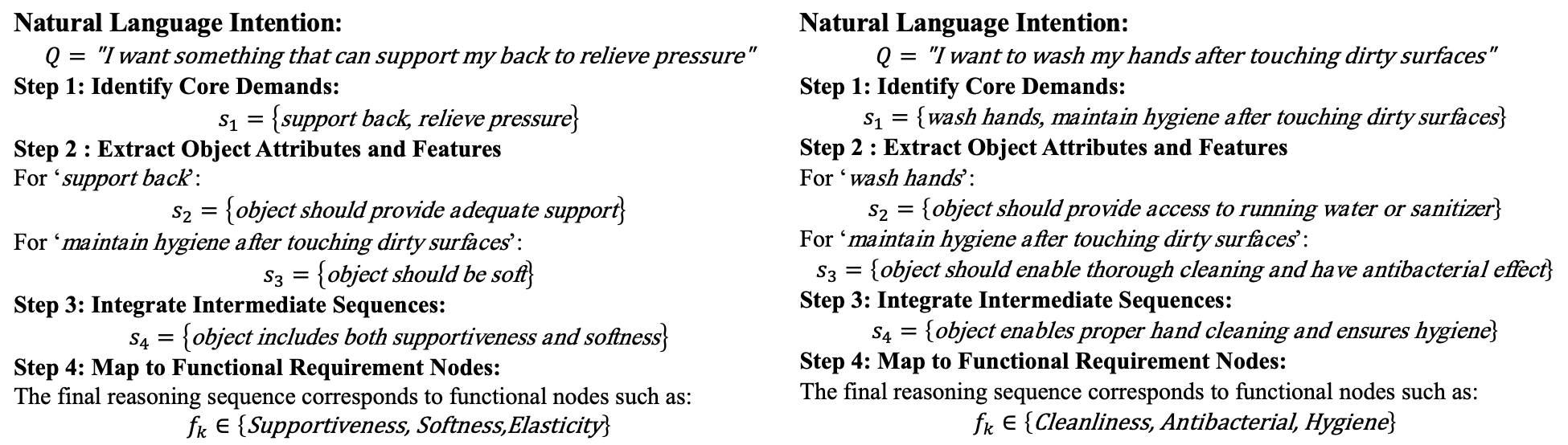}
  \caption{Step-wise Intent Causal Parsing Example.}
  \label{A1}
  \vspace{-15pt}
\end{figure*}

In Intent Causal Parsing (ICP), we decompose natural language intentions into functional requirements using a prompt template. The template is as follows: ‘Given a natural language intention, generate a step-by-step reasoning sequence to identify the functional requirements of the object. The reasoning process should first identify the core demands, then infer the required functional properties, and finally synthesize a complete set of functional requirements.’
For example, as shown in Figure~\ref{A1}, given the intention $Q=\text{‘I want something that can support my back to relieve pressure’}$, the model first identifies the core demands, producing $s_1=\{\text{support back}, \text{relieve pressure}\}$. It then infers the functional properties required by each core demand, resulting in $s_2=\{\text{requires supportiveness for back support}\}$ and $s_3=\{\text{requires softness for pressure relief}\}$. By integrating these intermediate steps, the model generates the final reasoning sequence $s_4=\{\text{required functions include supportiveness and softness}\}$. Finally, the reasoning path from the initial intention $s_0=Q$ to $s_4$ is mapped to the global functional requirement nodes. Each node $f_k\in\{f_1,\dots,f_K\}$ represents a distinct functional property. In this example, the reasoning sequence is mapped to functional nodes such as softness, elasticity, and supportiveness, forming the basis for causal graph construction.

\paragraph{T5-small Fine-tuning.}
We fine-tune T5-small on the training split only. Each training sample consists of an intention description as input and a step-wise functional reasoning sequence as output. The target sequence follows the same structure as the prompt template, including core demand identification, functional property inference, and final functional requirement summarization. The supervision is constructed from the training intent annotations and the corresponding functional cues extracted during the construction of the global functional requirement set. The target reasoning sequences are formatted according to the prompt structure without using validation or test annotations. The model is optimized with the standard sequence-to-sequence cross-entropy loss:
\begin{eqnarray}
\mathcal{L}_{\text{ICP}}=-\sum_{\ell=1}^{L}\log p(y_\ell \mid y_{<\ell}, Q),
\end{eqnarray}
where $y_\ell$ denotes the $\ell$-th token of the target reasoning sequence and $L$ is the output length. During inference, the fine-tuned T5-small generates the reasoning sequence conditioned on the input intent $Q$. To reduce error propagation from free-form generation, we do not directly use raw generated phrases for grounding. Instead, all generated reasoning steps are projected to the closed functional vocabulary $F$, producing a normalized functional distribution for subsequent causal grounding.

\paragraph{Construction of the Global Functional Requirement Set.}
To map parsed reasoning steps into a unified functional space, we construct a global functional requirement set $F=\{f_1,\dots,f_K\}$ from the training split. Different from object categories, functional requirements describe the intended usage or affordance implied by natural language intentions. For example, tables, desks, and counters may share the function of placement surface, while chairs, sofas, and beds may share the functions of seating, body support, and rest. Therefore, a compact functional vocabulary is necessary for bridging abstract intentions and concrete 3D object candidates.

Specifically, we construct the vocabulary from all intent annotations in the training split, which contains 35,850 intention-object pairs. Only the training split is used, and no validation or test descriptions are involved, thus avoiding information leakage. The construction process contains three steps. First, we collect all training intention descriptions and extract high-frequency action phrases and functional cues, including verbs, verb-object phrases, and usage-related expressions. Typical examples include ‘place my laptop’, ‘store my books’, ‘sit comfortably’, ‘throw away waste’, ‘write down notes’, and ‘display presentations’. Second, we merge synonymous or semantically overlapping functional cues into unified functional nodes. For instance, ‘put’, ‘place’, ‘lay out’, and ‘set down’ are merged into placement surface; ‘store’, ‘keep’, and ‘organize’ are merged into storage and organization; while ‘discard’, ‘dispose of’, and ‘throw away’ are merged into waste disposal. Third, we manually verify the resulting functional nodes to remove noisy, overly specific, or redundant entries, ensuring that each node represents a general and reusable functional requirement rather than an object-specific phrase.

After this process, the final vocabulary contains $K=40$ functional nodes, covering common indoor functional needs such as placement surface, work/study support, dining support, storage, organization, display, carrying, waste disposal, seating, rest, body support, writing, presentation, visual display, illumination, cleaning, bathing, cooking, liquid containment, wearing, decoration, hanging, spatial reference, and concealment.

\paragraph{Implementation Details.}
We first normalize all intention descriptions by lowercasing and removing noisy punctuation. For each functional node, we define a set of representative keywords and phrase patterns according to the extracted high-frequency functional cues. Each intention is matched against these patterns to estimate the coverage of each functional node. Since one intention may imply multiple functions, we allow a single intention to be assigned to multiple functional nodes. For example, the intention ‘I want to place my laptop and documents for a presentation’ can be associated with placement surface, work/study support, and presentation support simultaneously.

Formally, for each functional node $f_j$, we define a keyword set $\mathcal{K}_j=\{k_{j,1},\dots,k_{j,N_j}\}$, where $N_j$ denotes the number of representative keywords or phrases for $f_j$. Given an intention description $Q_i$, we assign a binary indicator $y_{ij}$ to indicate whether $Q_i$ contains functional cues related to $f_j$:
\begin{eqnarray}
y_{ij} =
\begin{cases}
1, & \text{if } Q_i \text{ matches any keyword or phrase in } \mathcal{K}_j, \\
0, & \text{otherwise}.
\end{cases}
\end{eqnarray}
The coverage frequency of each functional node is then computed as:
\begin{eqnarray}
c_j = \sum_{i=1}^{N} y_{ij},
\end{eqnarray}
where $N$ denotes the number of training intentions. Nodes with extremely low coverage or highly redundant semantics are further merged or removed during manual verification. Finally, each functional node is encoded into a text embedding and used as a global functional requirement prototype in ICP.

\subsubsection{More Details about Causal-Graph-Based Reasoning}
\label{Causal-Graph-Based Reasoning}

Our method performs reasoning on the constructed functional dependency graph to achieve final object selection. This process is not a one-shot matching operation, but an iterative refinement procedure that integrates linguistic functional reasoning and 3D visual evidence.

First, given the intent-level functional distribution $P(f_j \mid Q)$ obtained from Intent Causal Parsing, we initialize the object--function dependency $P(o_m \mid f_j)$ by measuring the functional compatibility between each candidate object $o_m$ and each functional requirement node $f_j$. The preliminary object score is computed by aggregating all function-mediated dependency paths:
\begin{eqnarray}
g_m = P(o_m \mid Q) = \sum_{j=1}^{K} P(o_m \mid f_j) \cdot P(f_j \mid Q),
\end{eqnarray}
where $P(f_j \mid Q)$ indicates the relevance of the $j$-th functional requirement to the input intent, and $P(o_m \mid f_j)$ denotes how well the $m$-th candidate object satisfies this functional requirement. This score provides an initial structured mapping from the abstract intent to candidate objects through intermediate functional requirements.

Next, the model refines this initial dependency graph through the Causal--Visual Feature Alignment (CVFA) module. Specifically, the functional requirement nodes are converted into function-aware representations:
\begin{eqnarray}
C_j = P(f_j \mid Q) \cdot \mathrm{Embed}(f_j),
\end{eqnarray}
where $\mathrm{Embed}(f_j)$ denotes the embedding of the $j$-th functional requirement. These function-aware representations are then fused with 3D candidate features through cross-attention:
\begin{eqnarray}
\tilde{V} = \mathrm{CrossAttn}(V, C),
\end{eqnarray}
where the candidate visual features $V$ serve as queries, and the function-aware representations $C$ serve as keys and values. In this way, each candidate object feature is updated according to intent-related functional cues.

To further refine the graph structure, we compute dependency-aware weights for object--function pairs:
\begin{eqnarray}
a_{mj} =
\frac{P(f_j \mid Q)P(o_m \mid f_j)}
{\sum_{m=1}^{M}\sum_{j=1}^{K} P(f_j \mid Q)P(o_m \mid f_j)}.
\end{eqnarray}
Based on these weights, the model encourages consistency only between functionally relevant object--requirement pairs:
\begin{eqnarray}
\mathcal{L}_{\mathrm{cons}} =
\sum_{m=1}^{M}\sum_{j=1}^{K}
a_{mj}\left(1-\cos(\tilde{v}_m, C_j)\right).
\end{eqnarray}
This dependency-weighted design avoids uniformly pulling all candidate objects toward all functional requirements. Instead, it strengthens reliable object--function associations while suppressing irrelevant or noisy paths.

After each refinement iteration, the object--function dependencies are updated according to the visual--functional consistency between $\tilde{v}_m$ and $C_j$. As a result, the functional dependency graph is progressively adjusted by visual evidence. For example, if a candidate object is visually consistent with the functional requirement ‘softness’ and this requirement is highly relevant to the input intent, the corresponding object--function dependency is reinforced. Conversely, visually inconsistent or weakly related dependencies are suppressed. The detailed iterative update mechanism is provided in Appendix~\ref{supp:iterative_details}.

Finally, after $N$ refinement iterations, the final object score is recomputed using the updated object--function dependencies:
\begin{eqnarray}
g_m = \sum_{j=1}^{K} P^{(N)}(o_m \mid f_j) \cdot P(f_j \mid Q),
\end{eqnarray}
where $P^{(N)}(o_m \mid f_j)$ denotes the refined object--function dependency after $N$ iterations. The final matched object is selected as:
\begin{eqnarray}
o^* = \arg\max_m g_m.
\end{eqnarray}
This iterative reasoning process enables the model to verify functional reasoning with 3D visual evidence, leading to more robust and interpretable object selection.

\subsubsection{Details of Iterative Reasoning Process}
\label{supp:iterative_details}

In the Causal--Visual Feature Alignment module, we employ an iterative mechanism to progressively refine candidate object features and update the functional dependency graph. This process aims to bridge the gap between high-level functional requirements and instance-level geometric--semantic evidence. Let $V^{(n)} \in \mathbb{R}^{M \times d}$ denote the visual features of the $M$ candidate objects at iteration $n$, where $V^{(0)}$ is extracted from the point cloud encoder, e.g., PointNet++. The function-aware feature bank $C=\{C_j\}_{j=1}^{K}\in\mathbb{R}^{K\times d}$ is constructed from the parsing distribution $P(f_j\mid Q)$, consistent with Eq.~(9) in the main paper.

For each iteration $n=1,\dots,N$, we first refine the object visual features in a function-guided manner and then recalibrate the object--function dependencies accordingly. Specifically, consistent with Eq.~(10) in the main paper, we employ the object features $V^{(n-1)}$ as queries and the function-aware features $C$ as keys and values to perform cross-attention:
\begin{equation}
\tilde{V}^{(n)} = \mathrm{CrossAttn}\big(V^{(n-1)}, C\big),
\end{equation}
where $\tilde{V}^{(n)}=\{\tilde{v}^{(n)}_m\}_{m=1}^{M}$ denotes the updated representations of candidate objects. We then set $V^{(n)} \leftarrow \tilde{V}^{(n)}$ for the subsequent dependency update.

With the refined visual features $V^{(n)}$, we update the functional dependency graph by recalibrating the object--function dependencies according to the current visual evidence. Specifically, we measure the alignment between each candidate object feature $v^{(n)}_m$ and each function-aware representation $C_j$, and normalize the dependencies over all candidate objects for each functional requirement node $f_j$:
\begin{equation}
P^{(n)}(o_m\mid f_j) =
\frac{\exp\big(\cos(v^{(n)}_m, C_j)\big)}
{\sum_{m'=1}^{M}\exp\big(\cos(v^{(n)}_{m'}, C_j)\big)}.
\end{equation}
This re-weighting mechanism strengthens dependency paths where the refined visual representation supports a functional requirement, while weakening irrelevant or visually inconsistent connections. Consistent with the pruning strategy in Eq.~(6) of the main paper, we optionally mask edges whose updated weights fall below the threshold $\theta$, i.e., $P^{(n)}(o_m\mid f_j)<\theta$, to maintain graph sparsity and suppress unreliable paths.

It is worth noting that Eq.~(5) in the main paper provides the initial object--function dependency based on text priors, where functional descriptions are matched with predicted category-name embeddings. In contrast, the iterative update described above dynamically recalibrates these dependencies using instance-level visual--functional consistency. Therefore, the graph is not fixed after initialization, but is progressively refined according to the visual evidence of the current 3D scene.

After $N$ iterations, we adopt a fixed-step inference strategy and terminate strictly at $n=N$. As analyzed in Fig.~\ref{hyper2}, setting $N=2$ provides a favorable trade-off between alignment depth and reasoning efficiency. Increasing the number of iterations beyond two brings marginal benefits and may homogenize object representations due to repeated fusion, leading to a slight performance drop.

Finally, the refined dependencies $P^{(N)}(o_m\mid f_j)$ are combined with the intent-level functional distribution $P(f_j\mid Q)$ to compute the final matching score for each candidate object:
\begin{equation}
g_m=\sum_{j=1}^{K} P(f_j\mid Q)\cdot P^{(N)}(o_m\mid f_j).
\end{equation}
The object with the highest score is selected as the final output:
\begin{equation}
o^* = \arg\max_{m} g_m.
\end{equation}
\subsection{More details of the experimental settings}
\subsubsection{Dataset}
\label{data}
Our main experiments are divided into two parts: the 3D Intention Grounding task and the 3D Visual Grounding task, providing a comprehensive evaluation of the effectiveness of our method.

\textbf{3D Intention Grounding (3D-IG) Task:} For this task, we adopt the official evaluation setting of the Intent3D dataset \cite{intent3d}. Intent3D is primarily derived from the ScanNet \cite{scannet} dataset and contains 44,990 intention texts across 1,042 ScanNet scenes. On average, each scene contains 61 instances and 43 texts, with each scene-object pair receiving 6 intention texts. The dataset includes 568 distinct verbs and 2,894 distinct nouns. Covering large-scale indoor 3D environments with paired natural language intention descriptions, Intent3D provides an effective benchmark for evaluating a model’s cross-modal reasoning ability between intention understanding and 3D object grounding. To ensure fair comparison with existing methods, we strictly follow the official data split, dividing the dataset into a training set (35,850 samples), a validation set (2,285 samples), and a test set (6,855 samples), with no overlapping scenes across splits.

\textbf{3D Visual Grounding Task:} We validate our method on the Nr3D and Sr3D datasets, both built upon ScanNet and widely used for 3D visual grounding. Nr3D \cite{Referit3d} contains 41,503 human-annotated referring expressions across 707 indoor scenes, where each description uniquely identifies a target object among distractors. Sr3D \cite{Referit3d} contains 83,572 machine-generated expressions based on predefined templates that focus on spatial relationships between objects. Both datasets provide ground-truth object masks. Performance is evaluated using accuracy, measuring the proportion of correctly localized target objects.

\subsubsection{Implementation Details}
\label{Imple}
 To ensure a fair comparison with existing methods, we adopt a common architecture for the foundational feature extractors, using PointNet++ \cite{pointnet++} for 3D point cloud feature extraction and RoBERTa-base \cite{roberta} for natural language intent encoding, consistent with baselines such as IntentNet. Our model is implemented in PyTorch and trained on a distributed system of four NVIDIA 3090 GPUs, each equipped with 24GB of memory, with a total batch size of 16. The intent causal parsing module employs a fine-tuned T5-small model \cite{t5}, with the hidden state dimension of the reasoning step vectors $d_s$ set to 512. We found that setting the number of reasoning steps $T$ in intent causal parsing to 4 yields the best results. The final visual and causal feature dimension $d$ is unified at 256. The entire model is optimized using the AdamW optimizer with a learning rate of $5 \times 10^{-5}$ and a weight decay of $1 \times 10^{-4}$, trained for a total of 120 epochs, taking approximately 40 hours. The total loss is a weighted sum, including grounding loss, causal pruning loss $\mathcal{L}_{\text{pruning}}$, and bidirectional consistency loss $\mathcal{L}_{\text{cons}}$, with the grounding loss weight set to 1.0 and both the causal pruning and bidirectional consistency losses set to 0.1. The causal pruning threshold $\theta$ is set to 0.2. For input settings, the number of point cloud tokens is fixed at 1024, and the maximum length of text tokens is set to 256. During training, our method runs end-to-end, parsing intent causal chains and constructing a causal graph to align functional requirements with object features, thereby achieving a unified causal reasoning process for object selection and localization.

\subsection{More Experiments Analysis}
\subsubsection{Validation on 3D Visual Grounding.}
\label{ScanRefer}
Table \ref{ScanRefertable} reports the 3D visual grounding results on ScanRefer \cite{Scanrefer} under the Unique and Multiple settings. Notably, Multiple queries account for the majority of samples ($\sim$81\%) and are generally more challenging due to higher instance ambiguity and stronger inter-object competition in cluttered indoor scenes. In this setting, our method achieves strong performance and remains competitive on both IoU thresholds, indicating that the proposed causal graph effectively captures attribute-level dependencies and helps disambiguate visually similar candidates. Meanwhile, our approach also performs well on the Unique subset ($\sim$19\%), suggesting that the causal modeling does not merely benefit crowded scenes but provides a consistent inductive bias for attribute reasoning and cross-modal alignment.
Compared with prior multimodal grounding approaches, the gains are largely attributed to explicit causal reasoning over functional/attribute cues rather than relying solely on implicit text–visual matching. By decomposing language into functional requirements and aligning them with geometric–semantic point-cloud features, our method can better distinguish objects with similar appearance but different functionality (e.g., different types of seating or storage furniture), which is particularly important for ScanRefer, where reference expressions often describe fine-grained attributes and relationships. In addition, the consistent improvements from IoU@0.25 to IoU@0.5 suggest that the proposed causal–visual alignment not only improves coarse retrieval but also enhances precise localization. Overall, these results demonstrate that CoCR generalizes beyond intent-centric benchmarks and provides a robust reasoning mechanism for 3D visual grounding in complex indoor environments.

\begin{table*}[t]
\centering
\caption{Performance Comparison on 3D Visual Grounding for ScanRefer DataSet.}
\label{ScanRefertable}
 \fontsize{12}{8}\selectfont
\resizebox{1.0\linewidth}{!}{
\begin{tabular}{l c c c c c c c}
\toprule
\toprule
\multirow{2}{*}{\textbf{Method}} & \multirow{2}{*}{\textbf{Venue}} & \multicolumn{2}{c}{\textbf{Unique ($\sim$19\%)}} & \multicolumn{2}{c}{\textbf{Multiple ($\sim$81\%)}} & \multicolumn{2}{c}{\textbf{Overall}} \\
\cmidrule(lr){3-4} \cmidrule(lr){5-6} \cmidrule(lr){7-8}
& & \textbf{0.25} & \textbf{0.5} & \textbf{0.25} & \textbf{0.5} & \textbf{0.25} & \textbf{0.5} \\
\midrule
ScanRefer \cite{Scanrefer} & ECCV20 & 67.6 & 46.2 & 32.1 & 21.3 & 40.0 & 26.1 \\

EDA \cite{EDA} & CVPR23 & 85.8 & 68.6 & 49.1 & 37.6 & 54.6 & 42.3 \\
VPP-Net \cite{VPPNet}& CVPR24 & 86.1 & 67.1 & 50.3 & 39.0 & 55.7 & 43.3 \\
M3DRef-CLIP \cite{M3DRef} & ICCV23 & - & 77.2 & - & 36.8 & - & 44.7 \\
G3-LQ \cite{G3LQ}& CVPR24 & 88.6 & 73.3 & 50.2 & 39.7 & 56.0 & 44.7 \\
MCLN \cite{MBVG} & ECCV24 & 86.9 & 72.7 & 52.0 & 40.8 & 57.2 & 45.5 \\
MA$^2$TransVG \cite{MAtrans}& CVPR24 & 86.3 & 74.1 & 53.8 & 41.4 & 57.9 & 45.7 \\
ConcreteNet \cite{ConcreteNet}& ECCV24 & 86.4 & 82.1 & 42.4 & 38.4 & 50.6 & 46.5 \\
D-LISA \cite{D-LISA}& NeurIPS24 & - & 75.5 & - & 40.0 & - & 46.9 \\
GPS \cite{GPS}& ECCV24 & - & 77.9 & - & 42.7 & - & 48.1 \\
Chat-Scene \cite{Chat-scene} & NeurIPS24 & 89.6 & \textbf{82.5} & 47.8 & 42.9 & 55.5 & 50.2 \\
PQ3D \cite{PQ3Q}& ECCV24 & 86.7 & 78.3 & 51.5 & 46.2 & 57.0 & 51.2 \\
Robin3D \cite{Robin3d}& ICCV25 & - & - & - & - & \textbf{60.8} & \textbf{55.1} \\
\midrule
Ours & - & \textbf{90.2} & 78.6 & \textbf{53.7} & \textbf{46.9} & 59.3 & 52.7 \\
\bottomrule
\bottomrule
\end{tabular}
}

\end{table*}

\subsubsection{Ablation of Components for 3D Visual Grounding Task.}
\label{Ab2}

To further assess the effectiveness of our method on the 3D Visual Grounding task, we performed an ablation study to evaluate the individual contributions of the three core modules in our framework, Intent Causal Parsing (ICP), Intent Causal Grounding (ICG), and Causal–Visual Feature Alignment (CVFA), summarized in Table \ref{Aab}. The first row reports the baseline performance. 
\begin{wraptable}{r}{0.5\textwidth}
\centering
\vspace{-10pt}
\caption{Ablation analysis of our proposed model on the Nr3D and Sr3D datasets.}
\resizebox{1.0\linewidth}{!}{
\begin{tabular}{lcccc}
\toprule
\toprule
\multirow{2}{*}{\textbf{Methods}} & \multicolumn{2}{c}{\textbf{Nr3D}} & \multicolumn{2}{c}{\textbf{Sr3D}} \\
\cmidrule(lr){2-3} \cmidrule(lr){4-5}
& \textbf{Overall} & \textbf{Hard} & \textbf{Overall} & \textbf{Hard} \\
\midrule
Baseline & 64.7 & 55.1  & 75.3 & 69.0 \\
+ ICP &  65.1 & 55.4 & 75.3 &  69.2 \\
+ ICG &  66.5 &  58.3 & 81.1 &  71.4\\
+ ICP + ICG & 66.4 & 58.9  & 80.9 & 72.4 \\
\midrule
+ ICP + ICG + CVFA & \textbf{67.1} & \textbf{59.8} & \textbf{81.6} & \textbf{73.7}  \\

\bottomrule
\bottomrule
\end{tabular}
}
\vspace{-10pt}
\label{Aab}
\end{wraptable} 
The second row shows results after incorporating ICP. As the 3D Visual Grounding task does not require explicit reasoning over textual information, ICP only performs step-by-step text decomposition without additional reasoning, resulting in minimal change in overall accuracy compared to the baseline.
The third row presents the performance with ICG, which shows a substantial improvement over the first two rows.
This demonstrates the benefit of explicitly modeling the causal relationships from text to objects, enabling the model better to understand functional attributes and their correspondence with visual features, thereby effectively distinguishing visually similar objects and achieving more precise localization in complex scenes.
The fourth row reports results with both ICP and ICG, yielding a modest improvement over ICG alone. Finally, the last row presents the performance after adding CVFA, which further enhances the results and confirms the necessity of causal–visual feature alignment. In summary, Table \ref{Aab} illustrates the individual effectiveness of each module as well as the performance gains from their combined integration, validating that our method is effective not only for 3D Intent Grounding but also for general 3D Visual Grounding tasks.

\begin{figure*}[t!]
  \centering
  \includegraphics[width=\textwidth]{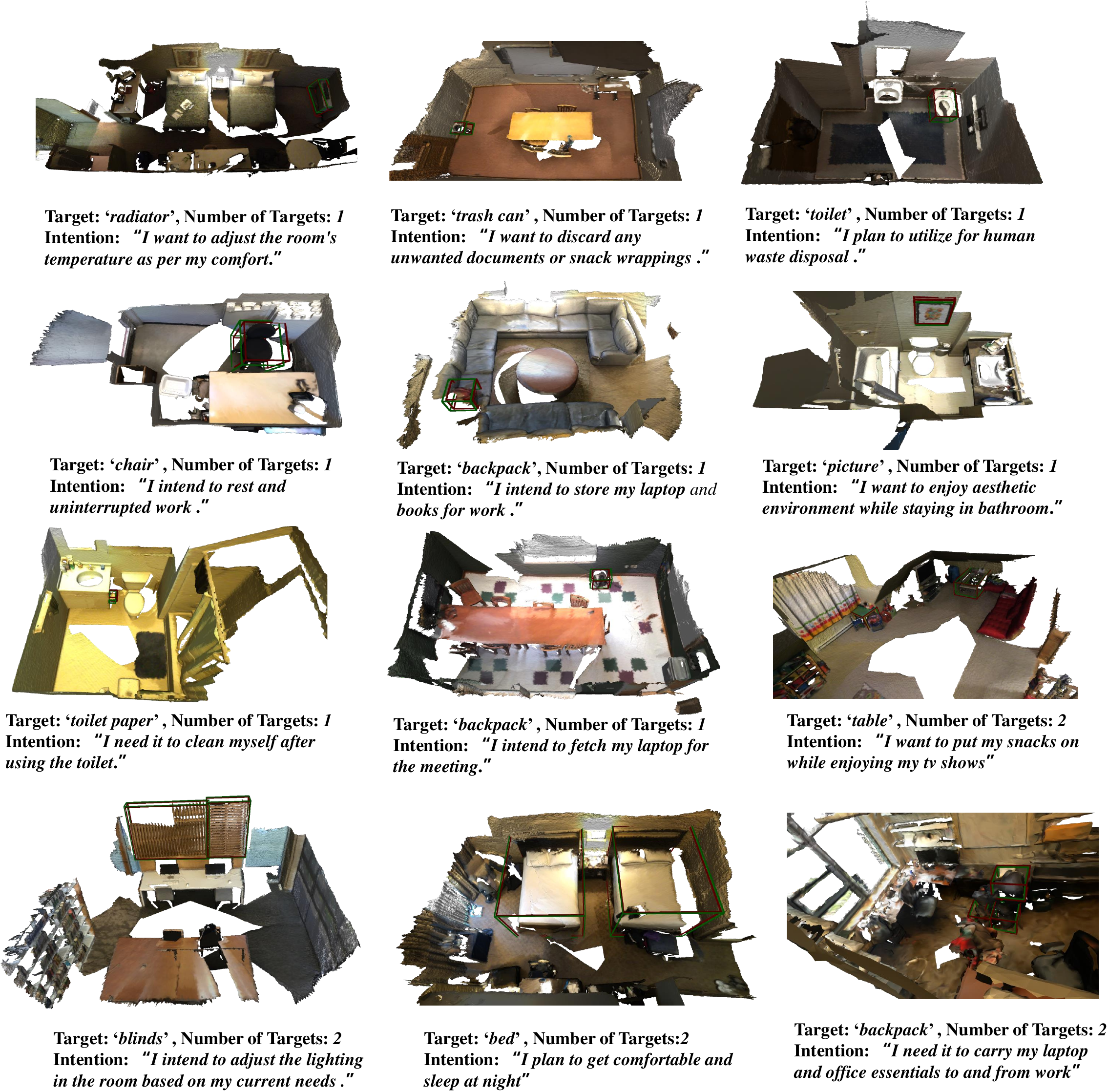}
  \caption{\textbf{More Visualization Results of Our Method on the Intent3D Dataset.} Green boxes indicate ground-truth annotations, while red boxes denote our detection results.  }
    \label{Mvis}
\end{figure*}

\subsubsection{Additional Comparison with Large Language Models}
\label{llm}
To evaluate the effectiveness of our method compared to directly using large language models (LLMs) for intention reasoning, we conducted experiments analogous to those for the IntentNet method, as summarized in Table \ref{LLM}. Specifically, for the LLM-based approach, we adopt a two-stage framework using the same detector as ours to predict bounding boxes and object categories. Given the intention and candidate boxes, the LLM analyzes the intent and selects the most relevant objects. The results are shown in Table \ref{LLM}. As observed, neither GPT nor Gemini achieves satisfactory detection performance. This is primarily due to the complexity of 3D scenes and the hallucination problem of LLMs when decoupled from 3D visual information.
Although LLMs like GPT-4 \cite{gpt4}, GPT-5 \cite{gpt5}, and Gemini \cite{gemini} possess strong textual understanding capabilities, they lack deep integration with 3D visual features. During intent analysis, they may produce “over-reasoning” hallucinations, such as misidentifying semantically related but irrelevant objects as targets. Moreover, they cannot leverage 3D spatial structures, object functionality, or other visual cues to verify their reasoning results. Consequently, the objects selected by LLMs often deviate from the true intent-compliant targets, leading to suboptimal overall detection performance.

\begin{table}[h]
\centering
\caption{Compare with the two-stage method based on Large Language Models.}
\label{LLM}
\resizebox{1.0\linewidth}{!}{
\begin{tabular}{cccccc}
\toprule
 \toprule
Method & Detector & Top1-acc@0.25 & Top1-acc@0.5 & Ap@0.25 & Ap@0.5 \\
 \midrule
Gemini \cite{gemini} & GroupFree & 39.21 & 29.34 & 12.17& 9.36 \\
GPT-4 \cite{gpt4} & GroupFree & 41.40 & 28.40 & 15.10 & 7.76 \\
GPT-5 \cite{gpt5} & GroupFree & 42.50 & 29.90 & 14.70 & 8.50 \\
IntentNet \cite{intent3d} & GroupFree & 58.34 & 40.83 & 41.90 & 25.36 \\
\midrule
 Ours & GroupFree & \textbf{60.71} & \textbf{43.24} & \textbf{42.65} & \textbf{27.33} \\
  \bottomrule
\bottomrule
\end{tabular}
}
\end{table}

\subsubsection{Analysis of the Number of Causal Graph Update Iterations $N$.}
\label{Causal Graph Update Iterations}
 \begin{wrapfigure}{r}{0.4\columnwidth}
  \centering
    \vspace{-15pt}
  \includegraphics[width=0.4\columnwidth]{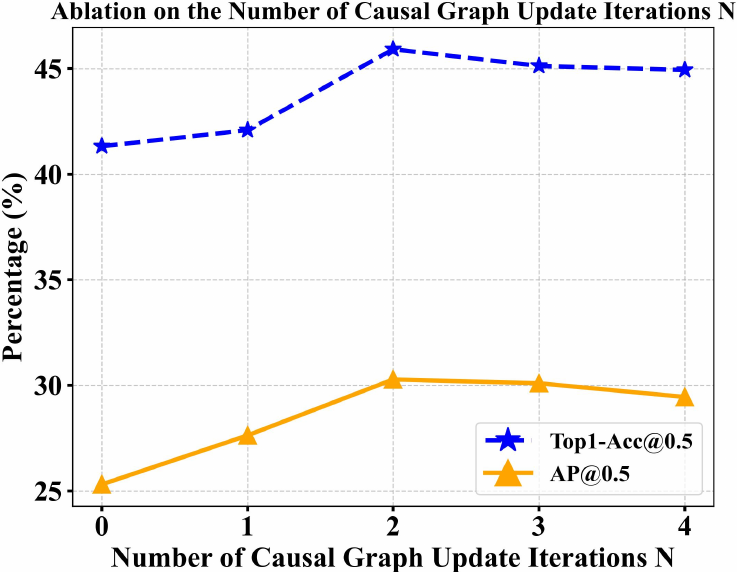}
  \vspace{-15pt}
  \caption{Analysis of the Number of Causal Graph Update Iterations $N$ in Causal-Visual Feature Alignment.}
    \label{hyper2}
\vspace{-10pt}
\end{wrapfigure}
\textbf{Analysis of the Number of Causal Graph Update Iterations $N$ in Causal-Visual Feature Alignment.} We investigate the impact of the causal graph update iterations $N$, a pivotal hyperparameter balancing alignment depth and efficiency. Specifically, insufficient iterations (e.g., $N=1$) fail to fully correct initial semantic mismatches between functional requirements and geometric features. Conversely, excessive iterations risk feature over-smoothing, where repeated message passing homogenizes distinct object representations and propagates noise. As shown in Fig. \ref{hyper2}, performance peaks at $N=2$, achieving the highest Top1-Acc@0.5 and AP@0.5. This indicates that two iterations effectively enforce robust bidirectional consistency without blurring object distinctiveness. Beyond this point, performance slightly declines due to over-smoothing and increased latency. For detailed algorithmic specifications and further analysis, please refer to Appendix \ref{supp:iterative_details}.

\subsubsection{Analysis of Intent Parsing and Grounding Accuracy}
\label{Intent Parsing}

\begin{wraptable}{r}{0.48\textwidth}
\centering
  \vspace{-20pt}
\caption{Evaluation of Intent Parsing and Grounding Accuracy.}
 \resizebox{1.0\linewidth}{!}{
\begin{tabular}{lcc}
\toprule
\toprule
\textbf{Method} & \textbf{Top-1 Acc (\%)} & \textbf{Top-5 Acc (\%)} \\
\midrule
GPT-5 \cite{gpt5} &   70.6         & 87.4          \\
IntentNet \cite{intent3d}       & 76.8   & 88.9          \\
Ours           & \textbf{83.2}                    & \textbf{91.3}                     \\
\bottomrule
\bottomrule
\end{tabular}
}
  \vspace{-10pt}
\label{intent}
\end{wraptable}

\textbf{Analysis of Intent Parsing and Grounding Accuracy.} To assess the effectiveness of our Intent Causal Parsing (ICP) and Intent Causal Grounding (ICG) modules, we conducted an evaluation on a sampled set of 1,000 intent texts, measuring the Top-1 and Top-5 accuracy of grounding parsed intents to the correct objects. As shown in Table \ref{intent}, our method outperforms both GPT-5 \cite{gpt5} and IntentNet \cite{intent3d}, achieving 83.2\% Top-1 accuracy and 91.3\% Top-5 accuracy. The results indicate that the step-wise decomposition of intents into functional requirements, coupled with explicit causal mapping to objects, significantly improves the model’s ability to correctly interpret and ground abstract intentions. Compared to IntentNet, which directly maps intent to objects without explicit causal reasoning, our method demonstrates superior precision in identifying the correct target object, validating the effectiveness of the causal parsing and grounding strategy.

\subsubsection{Additional visualization analysis.}
\label{Avis}
To further evaluate the effectiveness of our method, we conducted additional visualization experiments, as shown in Figure \ref{Mvis}, where the green boxes denote ground-truth objects and the red boxes represent the detections by our model. We present results across various scenes, including bathrooms, living rooms, and bedrooms, which involve diverse functional requirements. We also visualize both single-object and multi-object detection cases. As illustrated in the last row of Figure \ref{Mvis}, even when multiple target objects are present, our method can accurately parse the intent and localize the corresponding objects. Furthermore, when the target object is relatively small, for example, the ‘toilet paper’ in the first column of the third row, our method still achieves precise recognition. These results validate the robustness and effectiveness of our approach across diverse and complex scenarios.
In summary, the visualization experiments along three dimensions, multi-scene coverage, multi-object adaptation, and small-object recognition, fully demonstrate the effectiveness and robustness of the proposed method in complex 3D Intention Grounding, providing intuitive evidence of its practical applicability.

\begin{figure*}[h!]
  \centering
  \includegraphics[width=\textwidth]{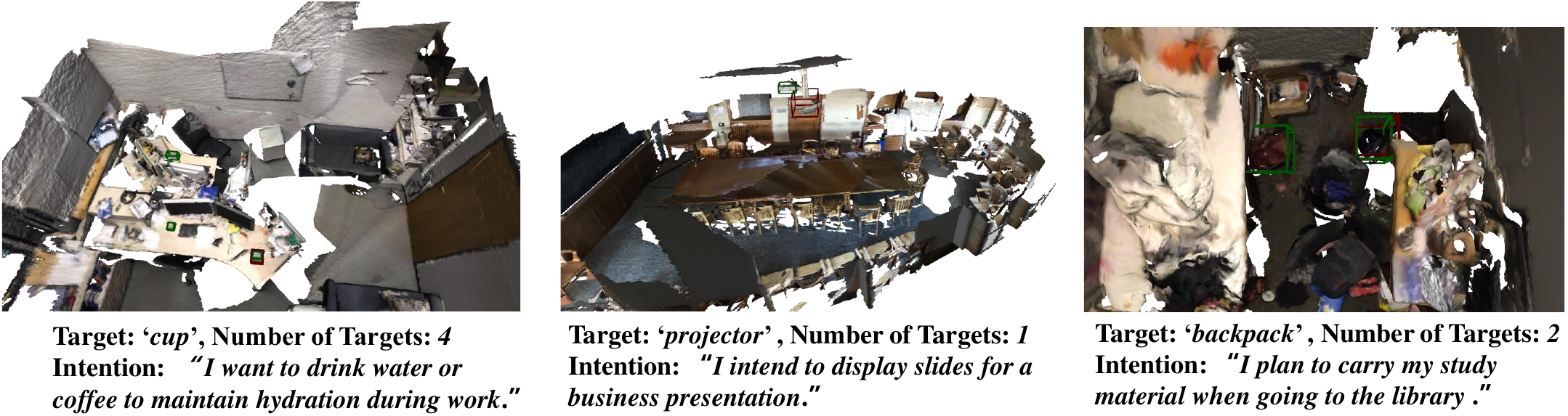}
  \caption{\textbf{Analysis of Failure Cases on the Intent3D Dataset.} Green boxes indicate ground-truth annotations, while red boxes denote our detection results. }
    \label{Fvis}
    
\end{figure*}

\subsubsection{Failure Cases and Model Limitations.}
\label{FCA}
Figure \ref{Fvis} presents several failure cases of our method, highlighting its current limitations. The first type of failure occurs when multiple target objects are present and their sizes are very small. In such cases, the model fails to detect all objects, leading to incomplete localization. This indicates that our method still struggles with densely packed small objects. The second type of failure involves detection box misalignment. In the illustrated scenario, the model parses the intent “I intend to display slides for a business presentation” and identifies the target object as a projector. However, due to inaccuracies in the 3D detection boxes, the final localization is incorrect. This case emphasizes that our method partially relies on the accuracy of candidate boxes: even if the intent parsing is correct, misaligned boxes can result in incorrect object localization. Improving candidate box precision is an important direction for future work. The third type of failure occurs in complex scenes with partial object occlusion, where the model detects only a subset of objects and fails to localize all targets in the scene. This demonstrates that in highly cluttered or occluded environments, our method may not achieve complete object localization.

\begin{figure*}[t!]
  \centering
  \includegraphics[width=\textwidth]{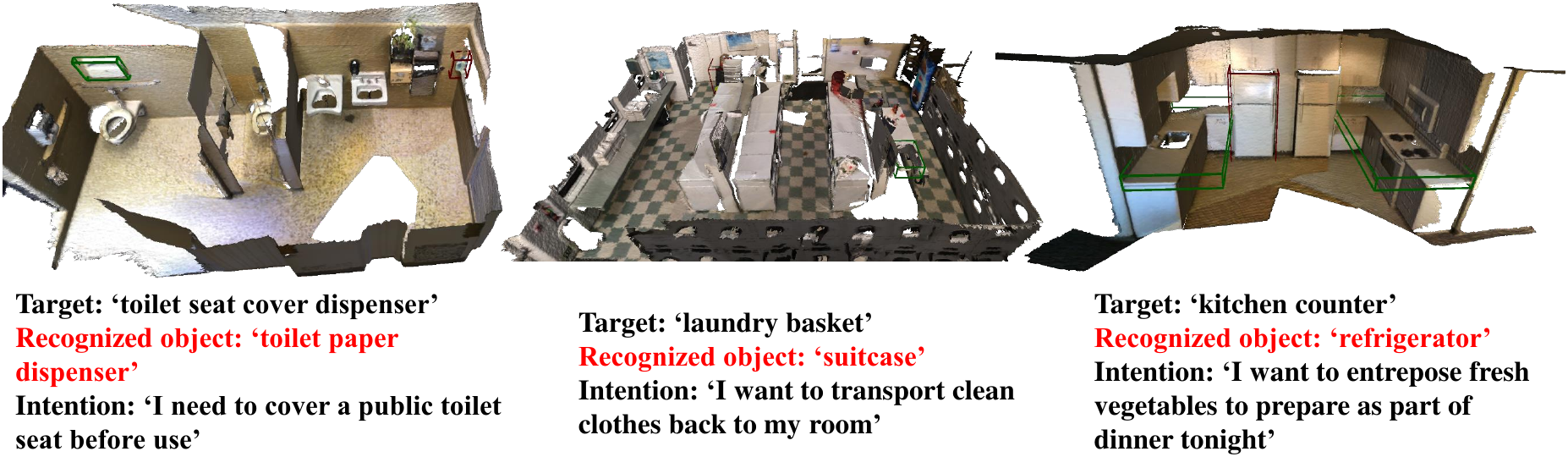}
  \caption{\textbf{Analysis of Intent Parsing Failure Cases.} Green boxes indicate ground-truth annotations, while red boxes denote our detection results.} 
    \label{Fvis-Intent}
    
\end{figure*}

Figure \ref{Fvis-Intent} illustrates several failure cases in intent parsing, highlighting the performance and limitations of our method when the parser makes errors. In the first case, the intent is to cover a toilet seat, and both the toilet paper and the toilet seat cover dispenser share overlapping functionality, specifically the “cover” function, with similar semantics. This causes an error in the intent decomposition, and the model ultimately recognizes the toilet paper dispenser instead of the correct target, the toilet seat cover dispenser. The second case similarly shows a recognition error due to overlapping functionality: the intent is “I want to transport clean clothes back to my room,” and during parsing, the model overemphasizes the “transport” function, resulting in the selection of a suitcase rather than the correct laundry basket. In the third case, the intent “I want to entrepose fresh vegetables to prepare as part of dinner tonight” contains the phrase “entrepose fresh vegetables.” The model focuses more on the storage function while neglecting the subsequent context “prepare as part of dinner tonight,” leading it to recognize the refrigerator used for storing vegetables instead of the kitchen counter used for preparing dinner.

Overall, the failure cases in Figure \ref{Fvis} provide valuable insights for future improvements, such as enhancing the detection of small objects, improving the accuracy of candidate boxes, and strengthening multi-object recognition under occlusion, thereby further boosting the robustness and practical applicability of our method. At the same time, the failure cases in Figure \ref{Fvis-Intent} highlight the limitations of our approach during the intent parsing stage. First, when different objects share overlapping functions or have similar semantics, the model is prone to confusion, leading to incorrect function–object matching. Second, the parser may overemphasize certain aspects of complex intents (e.g., “transport” or “storage”) while neglecting other critical contextual information, resulting in missed target objects. Finally, these issues are particularly pronounced for intents involving multi-step actions or composite functions, suggesting that future work should focus on improving the iterative reasoning of the causal graph and integrating contextual or multi-modal information to enhance the model’s robustness in scenarios with functionally similar intents or complex scenes.


\newpage
\section*{NeurIPS Paper Checklist}

\begin{enumerate}

\item {\bf Claims}
    \item[] Question: Do the main claims made in the abstract and introduction accurately reflect the paper's contributions and scope?
    \item[] Answer: \answerYes{}
    \item[] Justification: The abstract and introduction state the motivation, contributions, and scope of the proposed Chain-of-Causal Reasoning (CoCR) framework for 3D Intention Grounding. The claims are supported by the proposed intent parsing, functional causal dependency graph, causal--visual feature alignment, quantitative comparisons, ablation studies, efficiency analysis, and visualization results. We also clarify that the term ``causal'' refers to task-level functional causal dependencies rather than formal causal identification under structural causal models.
    \item[] Guidelines:
    \begin{itemize}
        \item The answer \answerNA{} means that the abstract and introduction do not include the claims made in the paper.
        \item The abstract and/or introduction should clearly state the claims made, including the contributions made in the paper and important assumptions and limitations. A \answerNo{} or \answerNA{} answer to this question will not be perceived well by the reviewers.
        \item The claims made should match theoretical and experimental results, and reflect how much the results can be expected to generalize to other settings.
        \item It is fine to include aspirational goals as motivation as long as it is clear that these goals are not attained by the paper.
    \end{itemize}

\item {\bf Limitations}
    \item[] Question: Does the paper discuss the limitations of the work performed by the authors?
    \item[] Answer: \answerYes{}
    \item[] Justification: The paper discusses limitations related to the task-level rather than formal SCM-based causal interpretation, the dependence on the T5-small intent parser, the coverage of the predefined functional requirement set, the quality of candidate object proposals, and robustness under ambiguous intentions, unseen functional requirements, and noisy or sparse point clouds. We also provide failure case analysis and discuss possible mitigation strategies such as confidence estimation, fallback rules, and abstention under uncertain predictions.
    \item[] Guidelines:
    \begin{itemize}
        \item The answer \answerNA{} means that the paper has no limitation while the answer \answerNo{} means that the paper has limitations, but those are not discussed in the paper.
        \item The authors are encouraged to create a separate ``Limitations'' section in their paper.
        \item The paper should point out any strong assumptions and how robust the results are to violations of these assumptions.
        \item The authors should reflect on the scope of the claims made.
        \item The authors should reflect on the factors that influence the performance of the approach.
        \item The authors should discuss the computational efficiency of the proposed algorithms and how they scale with dataset size.
        \item If applicable, the authors should discuss possible limitations of their approach to address problems of privacy and fairness.
        \item Reviewers will be specifically instructed to not penalize honesty concerning limitations.
    \end{itemize}

\item {\bf Theory assumptions and proofs}
    \item[] Question: For each theoretical result, does the paper provide the full set of assumptions and a complete (and correct) proof?
    \item[] Answer: \answerNA{}
    \item[] Justification: The paper does not include theoretical results, theorems, or formal proofs. CoCR is an empirically evaluated framework for 3D intention grounding. Although the method uses causal terminology, we do not claim formal causal identification, do-calculus, or structural causal model estimation.
    \item[] Guidelines:
    \begin{itemize}
        \item The answer \answerNA{} means that the paper does not include theoretical results.
        \item All the theorems, formulas, and proofs in the paper should be numbered and cross-referenced.
        \item All assumptions should be clearly stated or referenced in the statement of any theorems.
        \item The proofs can either appear in the main paper or the supplemental material.
        \item Any informal proof provided in the core of the paper should be complemented by formal proofs in appendix.
        \item Theorems and Lemmas that the proof relies upon should be properly referenced.
    \end{itemize}

\item {\bf Experimental result reproducibility}
    \item[] Question: Does the paper fully disclose all the information needed to reproduce the main experimental results of the paper to the extent that it affects the main claims and/or conclusions of the paper?
    \item[] Answer: \answerYes{}
    \item[] Justification: The paper describes the model architecture, Intent Causal Parsing, construction of the global functional requirement set, Intent Causal Grounding, causal pruning, Causal--Visual Feature Alignment, loss functions, datasets, official splits, evaluation metrics, hyperparameters, and implementation settings. Additional details about T5-small fine-tuning, functional vocabulary construction, iterative graph update, and experimental protocols are provided in the appendix.
    \item[] Guidelines:
    \begin{itemize}
        \item The answer \answerNA{} means that the paper does not include experiments.
        \item If the paper includes experiments, a \answerNo{} answer to this question will not be perceived well by the reviewers.
        \item If the contribution is a dataset and\slash or model, the authors should describe the steps taken to make their results reproducible or verifiable.
        \item Reproducibility can be accomplished in various ways.
    \end{itemize}

\item {\bf Open access to data and code}
    \item[] Question: Does the paper provide open access to the data and code, with sufficient instructions to faithfully reproduce the main experimental results, as described in supplemental material?
    \item[] Answer: \answerNo{}
    \item[] Justification: The datasets used in this paper, including Intent3D, ScanNet, Nr3D, Sr3D, and ScanRefer, are publicly available and properly cited. The code, parser implementation, functional vocabulary construction scripts, and trained models will be released upon acceptance with instructions for reproducing the main experiments.
    \item[] Guidelines:
    \begin{itemize}
        \item The answer \answerNA{} means that paper does not include experiments requiring code.
        \item Please see the NeurIPS code and data submission guidelines for more details.
        \item While we encourage the release of code and data, we understand that this might not be possible, so \answerNo{} is an acceptable answer.
        \item The instructions should contain the exact command and environment needed to run to reproduce the results.
        \item The authors should provide instructions on data access and preparation.
        \item The authors should provide scripts to reproduce all experimental results for the new proposed method and baselines.
        \item At submission time, to preserve anonymity, the authors should release anonymized versions if applicable.
    \end{itemize}

\item {\bf Experimental setting/details}
    \item[] Question: Does the paper specify all the training and test details necessary to understand the results?
    \item[] Answer: \answerYes{}
    \item[] Justification: The paper reports the datasets, official splits, evaluation metrics, point-cloud and text feature extractors, training epochs, batch size, hardware, point-cloud token number, maximum text length, functional vocabulary size, reasoning step number, causal pruning threshold, graph update iterations, and efficiency settings. More details about T5-small fine-tuning and functional requirement construction are included in the appendix.
    \item[] Guidelines:
    \begin{itemize}
        \item The answer \answerNA{} means that the paper does not include experiments.
        \item The experimental setting should be presented in the core of the paper to a level of detail that is necessary to appreciate the results and make sense of them.
        \item The full details can be provided either with the code, in appendix, or as supplemental material.
    \end{itemize}

\item {\bf Experiment statistical significance}
    \item[] Question: Does the paper report error bars suitably and correctly defined or other appropriate information about the statistical significance of the experiments?
    \item[] Answer: \answerNo{}
    \item[] Justification: The paper reports standard benchmark metrics following the official protocols of Intent3D and 3D visual grounding datasets, but does not include error bars, confidence intervals, or statistical significance tests due to the high computational cost of repeated full training runs. To support the conclusions, we provide comparisons with representative baselines, component-wise ablations, hyperparameter analyses, efficiency evaluation, visualization results, and failure case analysis.
    \item[] Guidelines:
    \begin{itemize}
        \item The answer \answerNA{} means that the paper does not include experiments.
        \item The authors should answer \answerYes{} if the results are accompanied by error bars, confidence intervals, or statistical significance tests.
        \item The factors of variability that the error bars are capturing should be clearly stated.
        \item The method for calculating the error bars should be explained.
        \item The assumptions made should be given.
        \item It should be clear whether the error bar is the standard deviation or the standard error of the mean.
        \item If error bars are reported in tables or plots, the authors should explain in the text how they were calculated.
    \end{itemize}

\item {\bf Experiments compute resources}
    \item[] Question: For each experiment, does the paper provide sufficient information on the computer resources needed to reproduce the experiments?
    \item[] Answer: \answerYes{}
    \item[] Justification: The paper reports the GPU type, GPU memory, number of GPUs, batch size, training time, inference latency per sample, peak inference memory usage, and number of parameters. These details are included in the implementation details and efficiency analysis.
    \item[] Guidelines:
    \begin{itemize}
        \item The answer \answerNA{} means that the paper does not include experiments.
        \item The paper should indicate the type of compute workers CPU or GPU, internal cluster, or cloud provider, including relevant memory and storage.
        \item The paper should provide the amount of compute required for each of the individual experimental runs as well as estimate the total compute.
        \item The paper should disclose whether the full research project required more compute than the experiments reported in the paper.
    \end{itemize}

\item {\bf Code of ethics}
    \item[] Question: Does the research conducted in the paper conform, in every respect, with the NeurIPS Code of Ethics?
    \item[] Answer: \answerYes{}
    \item[] Justification: The research uses existing public 3D scene datasets and benchmark protocols. It does not involve private user data, new human-subject experiments, or personally identifiable information. The study is conducted in accordance with the NeurIPS Code of Ethics.
    \item[] Guidelines:
    \begin{itemize}
        \item The answer \answerNA{} means that the authors have not reviewed the NeurIPS Code of Ethics.
        \item If the authors answer \answerNo{}, they should explain the special circumstances that require a deviation from the Code of Ethics.
        \item The authors should make sure to preserve anonymity.
    \end{itemize}

\item {\bf Broader impacts}
    \item[] Question: Does the paper discuss both potential positive societal impacts and negative societal impacts of the work performed?
    \item[] Answer: \answerYes{}
    \item[] Justification: The paper discusses the potential positive impact of improving intention-aware 3D scene understanding for embodied AI, human-computer interaction, assistive robotics, and indoor service agents. It also discusses possible negative impacts, including incorrect grounding decisions in safety-sensitive embodied scenarios and over-reliance on imperfect intent parsers. We include limitations, failure cases, and mitigation strategies to support responsible interpretation.
    \item[] Guidelines:
    \begin{itemize}
        \item The answer \answerNA{} means that there is no societal impact of the work performed.
        \item If the authors answer \answerNA{} or \answerNo{}, they should explain why their work has no societal impact or why the paper does not address societal impact.
        \item Examples of negative societal impacts include potential malicious or unintended uses, fairness considerations, privacy considerations, and security considerations.
        \item The conference expects that many papers will be foundational research and not tied to particular applications.
        \item The authors should consider possible harms that could arise when the technology is being used as intended and functioning correctly, when it gives incorrect results, and from misuse.
        \item If there are negative societal impacts, the authors could also discuss possible mitigation strategies.
    \end{itemize}

\item {\bf Safeguards}
    \item[] Question: Does the paper describe safeguards that have been put in place for responsible release of data or models that have a high risk for misuse?
    \item[] Answer: \answerNA{}
    \item[] Justification: The paper does not release high-risk pretrained generative models, scraped datasets, or systems with evident dual-use risks. The work focuses on 3D intention grounding using public benchmark datasets.
    \item[] Guidelines:
    \begin{itemize}
        \item The answer \answerNA{} means that the paper poses no such risks.
        \item Released models that have a high risk for misuse or dual-use should be released with necessary safeguards.
        \item Datasets that have been scraped from the Internet could pose safety risks.
        \item We encourage authors to take this into account and make a best faith effort.
    \end{itemize}

\item {\bf Licenses for existing assets}
    \item[] Question: Are the creators or original owners of assets used in the paper properly credited and are the license and terms of use explicitly mentioned and properly respected?
    \item[] Answer: \answerYes{}
    \item[] Justification: The paper cites the original works for all datasets, models, and baselines used, including Intent3D, ScanNet, Nr3D, Sr3D, ScanRefer, PointNet++, RoBERTa, T5-small, and the compared methods. We follow the official dataset protocols and terms of use.
    \item[] Guidelines:
    \begin{itemize}
        \item The answer \answerNA{} means that the paper does not use existing assets.
        \item The authors should cite the original paper that produced the code package or dataset.
        \item The authors should state which version of the asset is used and, if possible, include a URL.
        \item The name of the license should be included for each asset.
        \item For scraped data from a particular source, the copyright and terms of service should be provided.
        \item If assets are released, the license, copyright information, and terms of use in the package should be provided.
        \item For existing datasets that are re-packaged, both the original license and the license of the derived asset should be provided.
        \item If this information is not available online, the authors are encouraged to reach out to the asset's creators.
    \end{itemize}

\item {\bf New assets}
    \item[] Question: Are new assets introduced in the paper well documented and is the documentation provided alongside the assets?
    \item[] Answer: \answerNA{}
    \item[] Justification: This paper does not introduce a new dataset or benchmark asset. The method is evaluated on existing public datasets. The functional requirement vocabulary is constructed from the training split for model reasoning and will be documented with the released code upon acceptance.
    \item[] Guidelines:
    \begin{itemize}
        \item The answer \answerNA{} means that the paper does not release new assets.
        \item Researchers should communicate the details of the dataset\slash code\slash model as part of their submissions via structured templates.
        \item The paper should discuss whether and how consent was obtained from people whose asset is used.
        \item At submission time, remember to anonymize your assets if applicable.
    \end{itemize}

\item {\bf Crowdsourcing and research with human subjects}
    \item[] Question: For crowdsourcing experiments and research with human subjects, does the paper include the full text of instructions given to participants and screenshots, if applicable, as well as details about compensation?
    \item[] Answer: \answerNA{}
    \item[] Justification: This research does not involve new crowdsourcing experiments or new human-subject studies. It uses existing public benchmark datasets and their official annotations.
    \item[] Guidelines:
    \begin{itemize}
        \item The answer \answerNA{} means that the paper does not involve crowdsourcing nor research with human subjects.
        \item Including this information in the supplemental material is fine.
        \item According to the NeurIPS Code of Ethics, workers involved in data collection, curation, or other labor should be paid at least the minimum wage in the country of the data collector.
    \end{itemize}

\item {\bf Institutional review board (IRB) approvals or equivalent for research with human subjects}
    \item[] Question: Does the paper describe potential risks incurred by study participants, whether such risks were disclosed to the subjects, and whether IRB approvals were obtained?
    \item[] Answer: \answerNA{}
    \item[] Justification: The paper does not involve new studies with human subjects or new data collection from participants; therefore, IRB approval is not applicable.
    \item[] Guidelines:
    \begin{itemize}
        \item The answer \answerNA{} means that the paper does not involve crowdsourcing nor research with human subjects.
        \item Depending on the country in which research is conducted, IRB approval may be required for any human subjects research.
        \item We expect authors to adhere to the NeurIPS Code of Ethics and the guidelines for their institution.
        \item For initial submissions, do not include any information that would break anonymity.
    \end{itemize}

\item {\bf Declaration of LLM usage}
    \item[] Question: Does the paper describe the usage of LLMs if it is an important, original, or non-standard component of the core methods in this research?
    \item[] Answer: \answerYes{}
    \item[] Justification: The paper uses a fine-tuned T5-small model in the Intent Causal Parsing module to decompose natural-language intentions into step-wise functional reasoning sequences. The paper describes the prompt template, training split usage, input-output format, supervision construction, sequence-to-sequence loss, and the mapping from generated reasoning steps to the closed functional requirement vocabulary. The method does not rely on proprietary LLMs for inference, data generation, or experimental decision-making. Additional comparisons with LLM-based baselines are provided in the appendix.
    \item[] Guidelines:
    \begin{itemize}
        \item The answer \answerNA{} means that the core method development in this research does not involve LLMs as any important, original, or non-standard components.
        \item Please refer to our LLM policy in the NeurIPS handbook for what should or should not be described.
    \end{itemize}

\end{enumerate}

\end{document}